\documentclass[letterpaper, 10 pt, conference]{ieeeconf}
\usepackage{graphicx}
\usepackage{subcaption}
\usepackage{mwe}

\IEEEoverridecommandlockouts

\usepackage{mathrsfs}
\usepackage{amsmath}
\usepackage{amssymb}
\usepackage{amsfonts}
\usepackage{amsthm}
\usepackage{comment}
\usepackage{multirow}
\usepackage{cite}

\usepackage{enumerate}
\usepackage{cancel}
\usepackage{graphicx}
\usepackage{algorithm}
\usepackage[noend]{algpseudocode}
\usepackage[nameinlink,capitalise]{cleveref}
\usepackage{balance}
\usepackage{xcolor}
\usepackage{soul}
\usepackage[normalem]{ulem}

\usepackage{amssymb}

\DeclareMathOperator*{\argmin}{argmin}

\newtheorem{theorem}{Theorem}
\newtheorem{lemma}[]{Lemma}
\newtheorem{corollary}{Corollary}[theorem]
\newtheorem{definition}[]{Definition}

\newtheorem{problem}[]{Problem}
\newtheorem*{assumptions*}{Assumptions}
\newtheorem{assumption}[]{Assumption}
\newtheorem{proposition}[]{Proposition}
\newtheorem{example}{Example}

\usepackage{caption}
\usepackage{amsmath}
\allowdisplaybreaks

\newcommand{\gtrlesseqslant}{%
  \mathrel{\vcenter{\offinterlineskip
    \ialign{%
      \hfil$\mathsurround=0pt ##$\cr
      >\cr
      \noalign{\vskip-0.25ex}
      \leqslant\cr
    }%
  }}%
}

\title{\LARGE \bf Exploiting Trust for Resilient Hypothesis Testing with Malicious Robots}

\author{Matthew Cavorsi*, Orhan Eren Akgün*, Michal Yemini, Andrea Goldsmith, and Stephanie Gil
\thanks{(*Co-primary authors). M.~Cavorsi, O.~E.~Akgün, and S.~Gil are with the School of Engineering and Applied Sciences, Harvard University, USA:
        {\small mcavorsi@g.harvard.edu, erenakgun@g.harvard.edu, sgil@seas.harvard.edu}}%
\thanks{M.~Yemini is with the Faculty of Engineering, Bar-Ilan University, Israel: {\small michal.yemini@biu.ac.il}}%
\thanks{A.~Goldsmith is with the Department of Electrical and Computer Engineering, Princeton University, USA:
        {\small goldsmith@princeton.edu}}%
\thanks{The authors gratefully acknowledge partial support through AFOSR grant FA9550-22-1-0223 and  AFOSR award \#002484665.}%
\thanks{This paper was accepted in part for presentation at the 2023 IEEE International Conference on Robotics and Automation (ICRA) \cite{ICRA_hyp_2023}.}}

\begin{document}

\maketitle
\thispagestyle{empty}
\pagestyle{empty}

\begin{abstract}

We develop a resilient binary hypothesis testing framework for decision making in adversarial multi-robot crowdsensing tasks. This framework exploits stochastic trust observations between robots to arrive at tractable, resilient decision making at a centralized Fusion Center (FC) even when i) there exist malicious robots in the network and their number may be larger than the number of legitimate robots, and ii) the FC uses one-shot noisy measurements from all robots. We derive two algorithms to achieve this. The first is the Two Stage Approach (2SA) that estimates the legitimacy of robots based on received trust observations, and provably minimizes the probability of detection error in the worst-case malicious attack. Here, the proportion of malicious robots is known but arbitrary. For the case of an unknown proportion of malicious robots, we develop the Adversarial Generalized Likelihood Ratio Test (A-GLRT) that uses both the reported robot measurements and trust observations to estimate the trustworthiness of robots, their reporting strategy, and the correct hypothesis simultaneously. We exploit special problem structure to show that this approach remains computationally tractable despite several unknown problem parameters. We deploy both algorithms in a hardware experiment where a group of robots conducts crowdsensing of traffic conditions on a mock-up road network similar in spirit to Google Maps, subject to a Sybil attack. We extract the trust observations for each robot from actual communication signals which provide statistical information on the uniqueness of the sender. We show that even when the malicious robots are in the majority, the FC can reduce the probability of detection error to $30.5\%$ and $29\%$ for the 2SA and the A-GLRT respectively.

\end{abstract}

\IEEEpeerreviewmaketitle

\section{Introduction}
\begin{figure}[b!]
    \centering
    \includegraphics[scale=0.23]{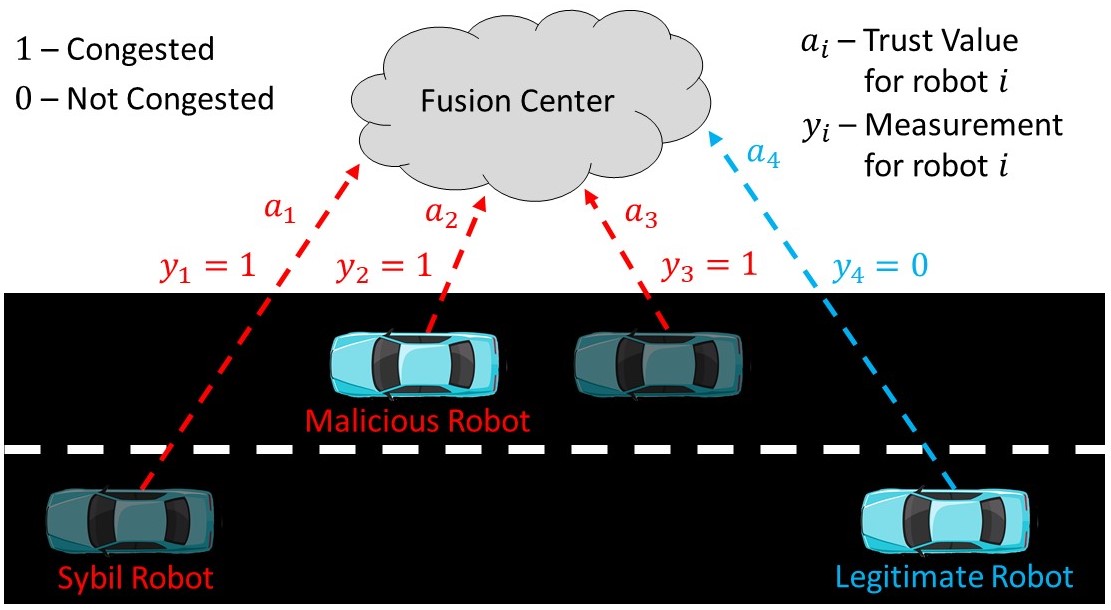}
    \caption[Problem setup for the work in binary hypothesis testing with malicious robots. The Fusion Center attempts to estimate traffic conditions on road segments while malicious robots perform a Sybil Attack to make roads look more congested.]{Malicious robots can perform a Sybil Attack to try to force a FC to incorrectly perceive traffic conditions on a road. The FC can aggregate measurements and trust values from robots to accurately estimate the true traffic condition of the road despite the attack.}
    \label{fig:map_spoof}
\end{figure}

We are interested in the problem where robots observe the environment and estimate the presence of an event of interest. Each robot relays its measurement to a \emph{Fusion Center} (FC) that makes an informed binary decision on the occurrence of the event. An unknown subset of the network are malicious robots whose goal is to increase the likelihood that the FC makes a wrong decision~\cite{kailkhura2014asymptotic, ren2018binary, althunibat2016countering, wu2018generalized}. This problem can be cast as an adversarial binary hypothesis testing problem, with relevance to a broad class of robotics tasks that rely on distributed sensing with possibly malicious or untrustworthy robots. For example, robots might perform coordinated coverage to maximize their ability to sense events of interest~\cite{robotTrustSchwager,xu2021novel,song2020care,talay2009task}, share target information for coordinated tracking~\cite{schlotfeldt2018resilient, ramachandran2020resilience,mitra2019resilient,laszka2015resilient}, or merge map information to provide a global understanding of the environment~\cite{blumenkamp2021emergence,mitchell2020gaussian, deng2021byz, wehbe2022probabilistically}. In crowdsensing tasks such as traffic prediction, a server may use GPS data to estimate if a particular roadway is congested or not~\cite{GoogleMaps} (see \cref{fig:map_spoof}). Unfortunately, this process is vulnerable to malicious robots~\cite{kailkhura2014asymptotic, althunibat2016countering}. For example, prior works have shown that a Sybil attack can cause crowdsensing applications like Google Maps to incorrectly perceive traffic conditions, resulting in erroneous reporting of traffic flows~\cite{jeske2013floating, wang2018ghost}.

The problem of binary adversarial hypothesis testing has been studied within the context of sensor networks~\cite{sandal2020reputation,  marano2008distributed, kailkhura2015distributed}. Many approaches use data, such as a history of measurements and hypothesis outcomes, to assess the trustworthiness of the robots~\cite{chen2008robust,nurellari2017secure, nurellari2016distributed, rawat2010collaborative}. For example, if a robot consistently disagrees with the final decision of the FC, then the FC can flag that robot as potentially adversarial. However, the success of these methods often hinges upon a crucial assumption that more than half of the network is legitimate. A growing body of work investigates additionally sensed quantities arising from the \emph{physicality of cyberphysical systems} such as multi-robot networks, to cross-validate and assess the trustworthiness of robots~\cite{trustandRobotsSycara,robotTrustSchwager,spoofResilientCoordinationusingFingerprints,securearray}. This could include using camera feeds, GPS signals, or even the signatures of received wireless communication signals, to acquire additional information regarding \emph{the trustworthiness} of the robots~\cite{AURO,securearray,CrowdVetting}. Importantly, this class of trust observations can often be obtained from a one-shot observation, independent of the transmitted measurement. The works in~\cite{yemini2021characterizing, yemini2022resilience,distOpt_arXiv} use trust observations to recover resilient consensus and distributed optimization even in the case where more than half of the network is malicious.  In this paper \emph{we wish to derive a framework for adversarial hypothesis testing that exploits stochastic trust observations to arrive at a similar level of resilience; whereby, a FC can conceivably reduce its probability of error, even in the one-shot scenario and where legitimate robots do not hold a majority in the network.}

We derive algorithms for achieving resilient hypothesis testing by exploiting stochastic trust observations between the FC and a group of robots participating in event detection. We derive a framework that exploits one-shot trust observations, hereafter called \emph{trust values}, over each link to arrive at \emph{tractable, closed-form solutions} when the majority of the network may be malicious and the strategy of the malicious robots is unknown -- a challenging and otherwise intractable problem to solve in the general case~\cite{soltanmohammadi2012decentralized}. 

For the case where an upper limit on the proportion of malicious robots is known, we develop the \emph{Two Stage Approach} (2SA). In the first stage this algorithm uses trust values to determine the most likely set of malicious robots, and then applies a Likelihood Ratio Test (LRT) only over trusted robots in the second stage. We show that this approach minimizes the error probability of the estimated hypothesis at the FC for a worst-case attack scenario. For the case where an upper bound on the proportion of malicious robots is unknown, we develop the \emph{Adversarial Generalized Likelihood Ratio Test} (A-GLRT) algorithm which uses both stochastic trust values and event measurements to jointly estimate the trustworthiness of each robot, the strategy of malicious robots, and the hypothesis of the event. Our A-GLRT algorithm is based upon a common approach for decision making with unknown parameters, the Generalized Likelihood Ratio Test (GLRT), which replaces the unknown parameters with their maximum likelihood estimates (MLE)~\cite{kay_2008}. We show that the addition of trust values allows us to decouple the trustworthiness estimation from the strategy of the adversaries, allowing us to calculate the exact MLE of unknown parameters in polynomial time, instead of approximating them as in previous works~\cite{soltanmohammadi2012decentralized,sun2016optimal}. Our simulation results show that the A-GLRT empirically yields a lower probability of error than the 2SA, but at the expense of higher computational cost. 

Finally we conduct a hardware experiment based on crowdsensing traffic conditions using a group of robots under a Sybil Attack. We show that the FC can recover a performance of $30.5\%$ and $29.0\%$ error, for the 2SA and A-GLRT respectively, even in the case where more than half of the robots are malicious.

This paper extends the results of the conference version \cite{ICRA_hyp_2023} in several distinct ways. First, it includes all the proofs that were excluded from the conference version due to space limitations. For the Two Stage Approach algorithm, additional analysis is provided regarding the probability of error for a fixed proportion of malicious robots as the number of robots in the network increases. In particular, we show that if the probability of trusting a legitimate robot is much higher than the probability of trusting a malicious robot, then the probability of error from using the Two Stage Approach will decay at least exponentially as the number of robots in the network increases. Additionally, we investigate the limiting behavior of the Two Stage Approach as the proportion of malicious robots becomes too high. We show that if there are too many malicious robots compared to legitimate robots in the network, the Two Stage Approach will resort to rejecting all information received, and choose a decision purely based on the probability of the event in question occurring. For the A-GLRT algorithm, we introduce two different modifications to the algorithm given any additional information. One modification utilizes knowledge of prior probabilities for legitimate and malicious robots, i.e., the probability any robot chosen at random will be legitimate or malicious. The other modification is helpful if there is a known upper bound on the number of malicious robots in the network.
Finally, we investigate the behavior of the A-GLRT as the \emph{quality} of the trust values improves, where a high quality trust value corresponds to a lower probability of misclassifying a legitimate robot as malicious, and vice versa. We show that as the quality becomes perfect, i.e., the trust values represent the true legitimacy of robots with probability $1$, the A-GLRT converge to the LRT using the measurements of the legitimate robots only.

\section{Related Works}

The system where a group of sensors detect an event locally and convey their binary measurements to the FC is well-studied in the literature \cite{kay_2008, varshney2012distributed}. The LRT minimizes the probability of error in the FC given that probability of false alarm and missed-detection for all sensors as well as the prior probability of the event is known by the FC \cite{kay_2008}. However, these distributed sensor networks are known to be susceptible to adversarial attacks as demonstrated by many previous works such as \cite{kailkhura2014asymptotic, ren2018binary,wu2018generalized}. In the presence of adversaries, the assumption of knowing the probability of false alarm and missed-detection of all sensors in the network doesn't hold anymore. Since the identities of the sensors in the network and the strategy of adversaries are unknown to the FC, the LRT cannot be employed in this setup. 

The problem of decision making with unknown parameters is known as composite hypothesis testing \cite{2001CDaE}. A common approach in composite hypothesis testing is to apply the GLRT which replaces the unknown parameters with their maximum likelihood estimates \cite{kay_2008}. The previous works in \cite{soltanmohammadi2012decentralized,sun2016optimal} approach the problem from this perspective by jointly determining the true hypothesis and estimating the unknown parameters in the system. The authors in \cite{soltanmohammadi2012decentralized} use an expectation-maximization algorithm to approximate the maximum likelihood estimates of the unknown parameters iteratively. At each iteration, the algorithm determines the identities of the sensors using the previously estimated false alarm and missed-detection probabilities. After that, the algorithm improves its estimation of these unknown probabilities by running a maximization step given the new identity estimation. After convergence, the LRT is applied using the estimated parameters. Similarly, the authors in \cite{sun2016optimal} propose a likelihood-based estimation algorithm for determining the identities of the sensors and their corresponding false-alarm and missed detection probabilities. Their iterative approach works similar to the expectation-maximization algorithm. The algorithm fixes all unknowns in the system but one, and they optimize over that free parameter. The algorithm they present improves the computational complexity over \cite{soltanmohammadi2012decentralized}, yet it still generates approximations to the maximum likelihood estimates. The A-GLRT algorithm we present is also based on the GLRT. It incorporates the trust observations into the GLRT framework. Moreover, our algorithm finds the exact maximum likelihood estimates instead of approximating them like previous works. 

Another common way to mitigate or anticipate the effect of adversaries in the network is to try to identify explicitly which robots are malicious. Previous works such as \cite{nurellari2016distributed, nurellari2017secure} identify malicious sensors using a reputation-based approach. In these approaches the FC compares the information received from each of the sensors with the final decision it arrives at over the course of several hypothesis tests. During this comparison, if the FC notices that certain sensors are consistently sending information that disagrees with the final decision, then those sensors can be flagged as potentially malicious. Then, the information from potentially malicious sensors can be given a smaller weight within the final decision scheme in order to favor using the information received from sources that have been mostly trustworthy. Other common ways to identify malicious sensors involve leveraging specific communication network structures. For example, the authors in \cite{hashlamoun2018audit} pair the sensors in groups of two. They implement an architecture where each sensor sends its information to the FC and also to the other sensor in its group. Then, each sensor also relays the information it received from its group member to the FC so the FC can examine the information for inconsistencies. If the FC finds a mismatch in the information it received, it can estimate that one of the two sensors within that particular group is not trustworthy. The authors in \cite{althunibat2016countering} assume that sensors report their measurements in a predefined order that is determined by the FC. Malicious sensors that report later in the reporting order can observe what sensors before them are reporting and choose to alter their value based on their observations. The FC detects which sensors are malicious by occasionally varying the reporting order and seeing how the performance of certain sensors changes. Then, given that the malicious sensors can be detected and their strategy can be discerned the authors create an optimal hypothesis decision rule that utilizes that information. In our work we look to similarly identify which sensors are potentially malicious in order to use that information to make a more informed decision. However, most of the related literature identifies malicious sensors by exploiting specific network structures or by referring to previous observations. This either restricts the network architectures that can be used or allows the FC to be susceptible for some time before it can develop a strategy to defend against the attack.

This paper considers a robotic sensor network where physical properties of the robot network may elucidate some additional information as to the trustworthiness of a particular robot. For example, in \cite{pippin2014trust,teacy2006travos} the robots physically interact with each other and each interaction has an expected outcome. The authors show that robots can determine the trustworthiness of neighboring robots by rating the outcome of their interaction is either successful or unsuccessful. Similarly, the authors in \cite{cheng2020there, cheng2021general} use Subjective Logic along with positive and negative evidence, which quantifies the observed satisfaction or violation of a particular property by a robot, to determine if a robot is trustworthy or not. The works in \cite{CrowdVetting,securearray,AURO} use physical properties of wireless transmissions to thwart Sybil attacks. They show that by analyzing the wireless profiles from incoming transmissions, certain transmissions can be determined to be malicious if their signal profiles are dishonest or too similar to another, hinting that the robot may be performing a spoofing attack. In all of these methods, it can be shown that the ability to confidently discern trust of a neighboring robot increases as more observations are made, but they can at least extract some useful information from even a single observations. Moreover, the authors in \cite{yemini2021characterizing} showed that since their method uses physical information that is independent of the information the robots transmit, the system can even handle scenarios where more than half of the robots in the network are malicious. We seek to leverage the benefits of these physical trust observations in order to improve the performance of a FC performing a binary hypothesis test in the presence of adversarial robots.

\section{Problem Formulation}
\label{sec:PF}

We consider a network of $N$ robots, where each robot is indexed by some $i \in \mathcal{N}$ and $\mathcal{N} = \{ 1, \dots, N \}$, that are deployed to sense an environment and determine if an event of interest has occurred. The event of interest is captured by the random variable $\Xi$, where $\Xi=1$ if the event has occurred and $\Xi=0$ otherwise. Each robot $i$ uses its sensed information to make a local decision about whether the event has happened or not, captured by the random variable $Y_i$, where its realization $y_i = 1$ if robot $i$ believes the event has happened and $y_i = 0$ otherwise. We denote by $\mathcal{H}_1$ the hypothesis that $\Xi = 1$ and $\mathcal{H}_0$ the hypothesis that $\Xi = 0$. Each robot forwards its local decision to a centralized fusion center (FC).

We are concerned with the scenario where not all robots are trustworthy, that is, some are \emph{malicious} and may manipulate the data that they send to the FC by flipping their measured bit with the goal of increasing the probability that the FC makes the wrong decision. We denote the set of malicious robots by $\mathcal{M} \subset \mathcal{N}$. The set of robots that are not malicious are termed \emph{legitimate robots}, denoted by $\mathcal{L} \subseteq \mathcal{N}$, where $\mathcal{L}\cup\mathcal{M}=\mathcal{N}$ and $\mathcal{L}\cap\mathcal{M}=\emptyset$. Additionally, we define the true trust vector, $\mathbf{t} \in \{0,1\}^N$, where $t_i = 1$ if $i \in \mathcal{L}$ and $t_i = 0$ if $i \in \mathcal{M}$. We note that the true trust vector is unknown by the FC, but it is defined for analytical purposes. We are interested in estimating this vector.

We assume the following behavioral models for legitimate and malicious robots:
\begin{definition}[Legitimate robot]
A \emph{legitimate robot} $i$ measures the event and sends its measurement $Y_i$ to the FC without altering it. We assume for each legitimate robot $i\in\mathcal{L}$, the measured bit $Y_i$ is subject to noise with the following false alarm and missed detection probabilities
\begin{equation}\label{eq:p_FA_MD_legitimate_sensor}
\begin{aligned}
P_{\text{FA},i} &= \Pr(Y_i=1|\Xi=0,t_i=1) = P_{\text{FA,L}}, \\
P_{\text{MD},i} &= \Pr(Y_i=0|\Xi=1,t_i=1) = P_{\text{MD,L}},
\end{aligned}
\end{equation} 
where $P_{\text{FA,L}} \in (0,0.5)$ and $P_{\text{MD,L}} \in (0,0.5)$ without loss of generality. We assume that all legitimate robots have homogeneous sensing capabilities, i.e., they have the same probability of false alarm and missed detection. Moreover, we assume that the measurement of a legitimate robot is independent of all other robots, and identically distributed given the true hypothesis. Finally, we also assume that $P_{\text{FA,L}}$ and $P_{\text{MD,L}}$ are known by the FC.
\end{definition}
\begin{definition}[Malicious robot]
A robot is said to be a \emph{malicious robot} if it can choose to alter its measurements before sending it to the FC.  We assume that a malicious robot $i\in\mathcal{M}$ can flip its measurement with probability $p_{\text{f}}\in [0,1]$ after making an observation, and that all malicious robots flip their bit with the same probability. Let $p_{\text{FA}, \text{M}}, p_{\text{MD}, \text{M}} \in [0,0.5)$ be the probability of false alarm and missed-detection of a malicious robot before altering the bit. We assume that all malicious robots have the same probability of false alarm and missed detection. The effective probabilities of false alarm and missed-detection of a malicious robot after altering the bit are given as:
\begin{flalign}\label{eq:p_FA_MD_malicious_sensor}
 P_{\text{FA,M}}&= \Pr(Y_i=1|\Xi=0,t_i=0) \\ &= (1-p_{\text{f}})\cdot p_{\text{FA,M}}+p_{\text{f}}\cdot (1-p_{\text{FA,M}}),\nonumber\\
P_{\text{MD,M}} &= \Pr(Y_i=0|\Xi=1,t_i=0) \\ &= (1-p_{\text{f}}) \cdot p_{\text{MD,M}}+p_{\text{f}}(1-p_{\text{MD,M}}).     \nonumber
\end{flalign}
We assume that a measurement coming from a malicious robot is independent of other measurements given the true hypothesis. This implies that malicious robots do not cooperate with each other. Furthermore, we assume that $p_{\text{FA}, \text{M}}$, $p_{\text{MD}, \text{M}}$, and the strategy of the malicious robots, which is the flipping probability $p_{\text{f}}$, are not known by the FC. This implies that the FC does not know $P_{\text{FA}, \text{M}}$ and $P_{\text{MD}, \text{M}}$ either.
\end{definition}
We use a common assumption in the literature which is that the measurements coming from malicious robots are i.i.d. (see \cite{ren2018binary, wu2018generalized, marano2008distributed, kailkhura2015distributed}). In addition to the measurements $Y_i$, we assume that each $Y_i$ is tagged with a \emph{trust value} $\alpha_i \in \mathbb{R}$. Specifically, we consider the class of problems where the FC can leverage the cyber-physical nature of the network to extract an estimation of trust about each communicating robot.
\begin{definition}[Trust Value $\alpha_i$]\label{def:trust_val}
A \emph{trust value} $\alpha_i$ is a stochastic variable  that captures information about the true legitimacy of a robot $i$. We denote the set of all possible trust values (aka sample space) by $\mathcal{A}$ and denote a realization for robot $i$ by $a_i$.
\end{definition}

\begin{assumption}\label{assumption:trust_val}
We assume that the set $\mathcal{A}$ is finite and that the trust value distributions are homogeneous across all the legitimate robots $i\in\mathcal{L}$. To this end, we denote the probability mass function of the trust values of robots by $p_{\alpha}(a|t)$.
We assume the probability mass functions are known or can be estimated by the FC.\footnote{\scriptsize{ Example of a trust value $\alpha_i$: One example of such trust values comes from the works in \cite{yemini2021characterizing, AURO, CrowdVetting}. In these works, the trust values $\alpha_i \in [0,1]$ are stochastic and are determined from physical properties of wireless transmissions. We use these trust values in our hardware experiment in \cref{sec:Results} where we discretize the sample space by letting $\mathcal{A} = \{0,1\}$ and find the probability mass functions to be $p_{\alpha}(a_i=1|t_i=1) = 0.8350$ and $p_{\alpha}(a_i=1|t_i=0) = 0.1691$. Other examples of observations can be found in \cite{trustandRobotsSycara, cheng2021general, peng2012agenttms}.}} We assume that the trust values are i.i.d. given the true legitimacy of the robot. 
Moreover, the trust values are assumed to be independent of the measurements, $Y_i$, and the true hypothesis. Finally, to omit trivial or noninformative cases, we assume that $p_{\alpha}(a| t = 0) \in (0,1)$, $p_{\alpha}(a| t = 1)\in(0,1)$, and $p_{\alpha}(a| t = 0) \neq p_{\alpha}(a| t = 1)$ for all $a\in\mathcal{A}$.
\end{assumption}

We do not impose any restrictions over the conditional probability distributions $p_{\alpha}(a | t = 1)$ and $p_{\alpha}(a | t = 0)$. However, for the trust values to be meaningful they should have different probability mass functions, i.e., $p_{\alpha}(a | t = 1) \neq p_{\alpha}(a | t = 0)$.
How distinguishable the two probability mass functions are is termed the \emph{quality} of the trust value, where a better quality corresponds to a larger distinction between the distributions $p_{\alpha}(a| t = 1)$ and $p_{\alpha}(a | t = 0)$. Based on these definitions, we provide the objective of the FC.

\subsection{The objective of the FC}

Denote the vector of all measurements with $\boldsymbol{Y}=(Y_{1},\ldots,Y_{N})$ and its realization $\boldsymbol{y}=(y_1,\ldots,y_N)$, and the vector of stochastic trust values by $\boldsymbol{\alpha}=(\alpha_{1},\ldots,\alpha_{N})$ and its realization by $\boldsymbol{a}=(a_{1},\ldots,a_{N})$. Let $\mathcal{D}_0$ and $\mathcal{D}_1$ be the decision regions at the FC. That is, $(\boldsymbol{a},\boldsymbol{y})\in\mathcal{D}_0$ if the FC chooses hypothesis $\mathcal{H}_0$ whenever it measures the  pair $(\boldsymbol{a},\boldsymbol{y})$. Similarly $(\boldsymbol{a},\boldsymbol{y})\in\mathcal{D}_1$ if the FC chooses hypothesis $\mathcal{H}_1$ whenever it measures the pair $(\boldsymbol{a},\boldsymbol{y})$. To simplify our notations we denote $\mathcal{D}:=\{\mathcal{D}_0,\mathcal{D}_1\}$.

Denote by $P_{\text{FA}}$ and $P_{\text{MD}}$ the false alarm and missed detection probabilities of the decision rule used by the FC, that is 
\begin{flalign}
&P_{\text{FA}}(\mathcal{D},\boldsymbol{t},P_{\text{FA,M}}) \nonumber\\
&\qquad= \sum_{(\boldsymbol{a},\boldsymbol{y})\in\mathcal{D}_1}\Pr(\boldsymbol{\alpha}=\boldsymbol{a},\boldsymbol{Y}=\boldsymbol{y}|\mathcal{H}_0,\boldsymbol{t},P_{\text{FA,M}}),\\
&P_{\text{MD}}(\mathcal{D},\boldsymbol{t},P_{\text{MD,M}}) \nonumber\\
&\qquad= \sum_{(\boldsymbol{a},\boldsymbol{y})\in\mathcal{D}_0}\Pr(\boldsymbol{\alpha}=\boldsymbol{a},\boldsymbol{Y}=\boldsymbol{y}|\mathcal{H}_1,\boldsymbol{t},P_{\text{MD,M}}).
\end{flalign}
Note that the false alarm and missed detection probabilities are affected by the strategy of the malicious robots, i.e., $P_{\text{FA,M}}$ and $P_{\text{MD,M}}$.

If the FC knows the true trust vector, i.e., the vector $\boldsymbol{t}$, and the probabilities $P_{\text{FA,M}}$ and $P_{\text{MD,M}}$, it could optimize the decision regions $\mathcal{D}_0$ and  $\mathcal{D}_1$ to minimize the expected error probability:
\begin{equation}
\begin{aligned}\label{eq:straightforward_obj_opt}
&P_{\text{e}}(\mathcal{D},\boldsymbol{t},P_{\text{FA,M}},P_{\text{MD,M}})= \\
&  \Pr(\Xi=0)P_{\text{FA}}(\mathcal{D},\boldsymbol{t},P_{\text{FA,M}})+\Pr(\Xi=1)P_{\text{MD}}(\mathcal{D},\boldsymbol{t},P_{\text{MD,M}}).
\end{aligned}
\end{equation}
In this case, the vector of trust values $\boldsymbol{\alpha}$ would not affect the optimal decision rule, and it would only depend on the vector of measurements $\boldsymbol{Y}$. 

However, there are two main obstacles to the optimization of the probability of error \eqref{eq:straightforward_obj_opt}, namely:
\begin{enumerate}
    \item The FC does not know the identity of the malicious robots, and thus it does not know the correct vector $\boldsymbol{t}$. Therefore, the FC needs to estimate the true trust vector, where the estimated trust vector is denoted by $\hat{\boldsymbol{t}}$.
    \item The FC does not know how the malicious robots alter their measurements before sending them. In our setup, this means that the FC does not know the values $P_{\text{FA,M}}$ and $P_{\text{MD,M}}$. Therefore the FC needs to estimate $P_{\text{FA,M}}$ and $P_{\text{MD,M}}$, where the estimates are denoted by $\hat{P}_{\text{FA,M}}$ and $\hat{P}_{\text{MD,M}}$, respectively.
\end{enumerate}

The FC needs to make a decision with these unknown parameters which is known as the composite hypothesis testing problem. Since the minimization of \eqref{eq:straightforward_obj_opt} is not tractable, we explore different ways to circumvent this issue. One way is to start by estimating the legitimacy of the robots using trust values only and assuming that the upper bound on the number of malicious robots in the network is known in order to make \eqref{eq:straightforward_obj_opt} tractable. Then, we can ignore the measurements from robots deemed to be malicious and choose the decision regions $\mathcal{D}_0$ and $\mathcal{D}_1$ using the measurements from the remaining robots. This approach leads us to the formulation in \cref{prob:prob1}.

\begin{problem} \label{prob:prob1}
Assume that the FC first estimates the identities of the robots in the network, i.e., it determines $\hat{\boldsymbol{t}}$, solely using the vector of trust values $\boldsymbol{\alpha}$. Then, the FC makes a decision about the hypothesis using only the vector of measurements $\mathbf{Y}$, from robots it identifies as legitimate. Given an upper bound $\overline{m}$ on the proportion of malicious robots in the network, we wish to determine a strategy for the FC that minimizes the following worst-case scenario under these assumptions:
\begin{flalign}
   \min_{\mathcal{D}}\max_{P_{\text{FA,M}},P_{\text{MD,M}},\boldsymbol{t}:\sum_{i\in\mathcal{N}}t_i\leq \overline{m}N} P_{\text{e}}(\mathcal{D},\boldsymbol{t},P_{\text{\emph{FA,M}}},P_{\text{\emph{MD,M}}}).
\end{flalign}

\end{problem}

The definition in \cref{prob:prob1} requires an approach that estimates the trustworthiness of a robot $i$ using only the trust value $a_i$ associated with that robot while assuming a known upper bound on the proportion of malicious robots. However, it is natural to seek additional information about the trustworthiness of the robots that can be obtained from the random measurement vector $\boldsymbol{y}$. Following this intuition, we seek a decision rule that estimates the unknown parameters in the system which are $\boldsymbol{t}$, $P_{\text{FA,M}}$, and $P_{\text{MD,M}}$ as well as the hypothesis $\mathcal{H}_0$ or $\mathcal{H}_1$ jointly, without requiring any known upper bound on the proportion of malicious robots. A common approach to hypothesis testing with unknown parameters is to use the generalized likelihood ratio test \cite{kay_2008}, that is
\begin{flalign}\label{eq:GLRT_ML_decision}
\frac{p(\boldsymbol{z};\hat{\theta}_1,\mathcal{H}_1)}{p(\boldsymbol{z};\hat{\theta}_0,\mathcal{H}_0)}\underset{\mathcal{H}_{0}}{\overset{\mathcal{H}_{1}}{\gtrlesseqslant}} \:\frac{\Pr(\Xi=0)}{\Pr(\Xi=1)} \triangleq {\gamma}_{\text{GLRT}},
\end{flalign}
where $\hat{\theta}_1$ is the maximum likelihood estimator (MLE) of the unknown parameter $\theta_1$ assuming $\Xi=1$ and $\hat{\theta}_0$ is the MLE of $\theta_0$ assuming $\Xi=0$. For our problem, $\boldsymbol{z}=(\boldsymbol{a},\boldsymbol{y})$, $\theta_1 = (\boldsymbol{t},P_{\text{MD,M}})$, and
$\theta_0 = (\boldsymbol{t},P_{\text{FA,M}}),$ which results in the following formulation of the test 
\begin{flalign}\label{eq:GLRT_ML_decision_joint}
\frac{\max_{\boldsymbol{t}\in\{0,1\}^N,P_{\text{MD,M}}\in[0,1]}\Pr(\boldsymbol{a},\boldsymbol{y}|\mathcal{H}_1,\boldsymbol{t},P_{\text{MD,M}})}{\max_{\boldsymbol{t}\in\{0,1\}^N,P_{\text{FA,M}}\in[0,1]}\Pr(\boldsymbol{a},\boldsymbol{y}|\mathcal{H}_0,\boldsymbol{t},P_{\text{FA,M}})}\underset{\mathcal{H}_{0}}{\overset{\mathcal{H}_{1}}{\begin{smallmatrix}>\\\leqslant\end{smallmatrix}}} \gamma_{\text{GLRT}}.
\end{flalign}

Note that in this setup the vector $\boldsymbol{t}$ is a parameter, thus, we do not make any prior assumption on its distribution. Calculating the MLE in the numerator and denominator in \eqref{eq:GLRT_ML_decision_joint} is not trivial since the unknown $\boldsymbol{t}$ is a discrete multidimensional variable while $P_{\text{MD,M}}$ and $P_{\text{FA,M}}$ are continuous variables. Doing this in a tractable way leads us to the formulation in \cref{prob:prob2}.

\begin{problem} \label{prob:prob2}
Find a computationally tractable algorithm that calculates the GLRT given in \eqref{eq:GLRT_ML_decision_joint}.
\end{problem}

In the next sections we propose two different approaches: one approach to solve \cref{prob:prob1} and another to solve \cref{prob:prob2}. Then, we investigate the performance of both methods in Section~\ref{sec:Results}, and conclude the paper in Section~\ref{sec:conclusion}.

\section{Two Stage Approach}
The first approach, called the Two Stage Approach, finds the optimum decision rule that solves \cref{prob:prob1}.

\subsection{Two Stage Approach Algorithm}
\label{sec:2SA}

In this section we present an intuitive approach where we separate the detection scheme into two stages where 1) a decision is made about the trustworthiness of each individual robot $i$ based on the received value $\alpha_i$, and then 2) only the measurements $Y_i$ from robots that are trusted are used to choose $\mathcal{H}_0$ or $\mathcal{H}_1$.

\paragraph{Detection of Trustworthy Robots}
We utilize the Likelihood Ratio Test (LRT) to detect \textit{legitimate} robots. This test is guaranteed to have minimal missed detection probability (i.e., detecting a legitimate robot as malicious) for a given false alarm probability (i.e., detecting a malicious robot as legitimate) \cite[Chapter 3]{kay_2008}.

The FC decides which robots to trust using the LRT decision rule
\begin{flalign}\label{eq:classification_decision}
\frac{p_{\alpha}(a_i|t_i=1)}{p_{\alpha}(a_i|t_i=0)}\underset{\hat{t}_i = 0}{\overset{\hat{t}_i = 1}{\gtrless}} \gamma_t,
\end{flalign}
where $\gamma_t$ is a threshold value that we wish to optimize. Note that when $\gamma_t=1$ \eqref{eq:classification_decision} is equivalent to a maximum likelihood detection.

The FC decides who to trust and stores it in the vector $\mathbf{\hat{t}}$, where $\hat{t}_i = 1$ if the FC chooses to trust the robot, and $\hat{t}_i = 0$ otherwise. In the case of equality, a random decision is made where the FC chooses $\hat{t}_i = 1$ with probability $p_t$ and the FC chooses $\hat{t}_i = 0$ with probability $1-p_t$, where $p_t$ is another parameter to be optimized. 
This leads to the following trust probabilities, where $P_{\text{trust,L}}(\gamma_t,p_t)$ is the probability of trusting a legitimate robot, and $P_{\text{trust,M}}(\gamma_t,p_t)$ is the probability of trusting a malicious robot:
\begin{equation}
\begin{aligned}
     P_{\text{trust,L}}(\gamma_t,p_t)&= \Pr\left(\frac{p_{\alpha}(a_i|t_i=1)}{p_{\alpha}(a_i|t_i=0)}>\gamma_t \bigg| t_i=1\right)\\
     &+p_t  \Pr\left(\frac{p_{\alpha}(a_i|t_i=1)}{p_{\alpha}(a_i|t_i=0)}=\gamma_t \bigg| t_i=1\right),\\
    P_{\text{trust,M}}(\gamma_t,p_t)&= \Pr\left(\frac{p_{\alpha}(a_i|t_i=1)}{p_{\alpha}(a_i|t_i=0)}>\gamma_t \bigg| t_i=0\right)\\
     &+p_t  \Pr\left(\frac{p_{\alpha}(a_i|t_i=1)}{p_{\alpha}(a_i|t_i=0)}=\gamma_t \bigg| t_i=0\right).
     \label{eq:trust_prob_two_stage}
\end{aligned}
\end{equation}
The error probability $P_{\text{e}}(\mathcal{D},\boldsymbol{t},P_{\text{FA,M}},P_{\text{MD,M}})$ at the FC in \eqref{eq:straightforward_obj_opt} is affected by the trustworthiness classification. That is, if a legitimate robot~$i$ is classified as malicious the FC discards its measurement $Y_i$, which increases the error probability since fewer measurements are used in the FC decision making. On the other hand, if a malicious robot is classified as legitimate it can increase the error probability by sending falsified measurements to the FC. For that reason, we look to optimize the trustworthiness classification to balance these two conflicting scenarios. Determining the best $\gamma_t$ and $p_t$ to minimize the overall error probability of the hypothesis detection by the FC is the main focus of this section.

\paragraph{Detecting the Event $\Xi$} 
To determine a hypothesis $\mathcal{H}$ on the event $\Xi$, the FC only considers the measurements it receives from robots that it classifies as legitimate in the first stage, i.e., $i: \hat{t}_i=1$. Equivalently, the FC discards all the received measurements of robots it classifies as malicious. 
Then, the FC uses the following decision rule:
\begin{flalign}\label{eq:detection_p_e_FC_legitimate_assumption}
\frac{\prod_{\{i:\hat{t}_i=1\}}P_{\text{MD},\text{L}}^{1-y_i}(1-P_{\text{MD},\text{L}})^{y_i}}{\prod_{\{i:\hat{t}_i=1\}}(1-P_{\text{FA},\text{L}})^{1-y_i}P_{\text{FA},\text{L}}^{y_i}}\underset{\mathcal{H}_{0}}{\overset{\mathcal{H}_{1}}{\begin{smallmatrix}\geqslant\\<\end{smallmatrix}}}\frac{\Pr(\Xi=0)}{\Pr(\Xi=1)} = \exp(\gamma_{\text{TS}}),
\end{flalign}
where $\exp(\gamma_{\text{TS}})$ is the exponential function with respect to $\gamma_{\text{TS}}$, and it is a constant decision threshold. We set $\frac{\Pr(\Xi=0)}{\Pr(\Xi=1)} = \exp(\gamma_{\text{TS}})$ so that when we take the logarithm in later expressions we can express the resultant decision threshold as $\gamma_{\text{TS}}$ for ease of exposition. This decision rule is commonly used in standard binary hypothesis testing problems where no malicious robots are present, and will be referred to as the \emph{standard binary hypothesis decision rule}. The standard binary hypothesis decision rule is optimal in a system with no malicious robots, i.e., ${\cal M}=\emptyset$, and thus we attempt to approximate the standard binary hypothesis decision rule by first removing information from all robots deemed to be malicious. However, since there may be detection errors in the first stage which classifies legitimate and malicious robots, the threshold $\gamma_t$ and tie-break probability $p_t$ should  balance the need to exclude malicious robots from participating in the test \eqref{eq:detection_p_e_FC_legitimate_assumption} with the need to allow legitimate robots to participate in the test \eqref{eq:detection_p_e_FC_legitimate_assumption} and contribute their truthful measurements to decrease the probability of error resulting from \eqref{eq:detection_p_e_FC_legitimate_assumption}. In what follows we show how to optimize the threshold $\gamma_t$ and tie-break probability $p_t$ by first computing the probability of error of the FC using the Two Stage Approach.

Recalling the Neyman-Pearson Lemma \cite{kay_2008}, we have that \eqref{eq:classification_decision}
minimizes the missed detection probability for a desired false alarm probability of misclassifying robots. This false alarm probability dictates the value of the threshold $\gamma_t$. After the FC discards robot measurements that it does not trust, the decision rule \eqref{eq:detection_p_e_FC_legitimate_assumption} leads to the following false alarm and missed detection error probabilities,
 \begin{equation}
 \begin{aligned}
    &P_{\text{FA}}(\gamma_t,p_t,\mathbf{t},P_{\text{FA,M}}) \\
    & =\Pr\Big( \sum_{i=1}^N \hat{t}_i [w_{1,\text{L}} y_i - w_{0,\text{L}}(1-y_i)] \geq \gamma_{\text{TS}} \\
     &\hspace{4.5cm} \Big| \mathcal{H}_0, \gamma_t,p_t,\mathbf{t},P_{\text{FA,M}} \Big), \\ 
    &P_{\text{MD}}(\gamma_t,p_t,\mathbf{t},P_{\text{MD,M}}) \\
    &= \Pr\Big( \sum_{i=1}^N \hat{t}_i [w_{1,\text{L}} y_i - w_{0,\text{L}}(1-y_i)] < \gamma_{\text{TS}} \\
    &\hspace{4.5cm} \Big| \mathcal{H}_1 ,\gamma_t,p_t,\mathbf{t},P_{\text{MD,M}}\Big),
        \label{eq:fa_md_total}
    \end{aligned}
    \end{equation}
where
\begin{equation}
    w_{1,\text{L}} = \log\left( \frac{1 - P_{\text{MD},\text{L}}}{P_{\text{FA},\text{L}}} \right), \quad w_{0,\text{L}} = \log\left( \frac{1 - P_{\text{FA},\text{L}}}{P_{\text{MD},\text{L}}} \right).
    \label{eq:w1_w0}
\end{equation}
Consequently, the overall error probability at the FC is:
\begin{equation}
\begin{aligned}
    &P_{\text{e}}(\gamma_t,p_t,\mathbf{t},P_{\text{FA,M}},P_{\text{MD,M}})  \\ 
    &= \Pr(\Xi=0) P_{\text{FA}}(\gamma_t,p_t,\mathbf{t},P_{\text{FA,M}}) \\ 
    &+ \Pr(\Xi=1) P_{\text{MD}}(\gamma_t,p_t,\mathbf{t},P_{\text{MD,M})}.
    \label{eq:error_prob_2stage}
\end{aligned}
\end{equation}
   
We seek to minimize the probability of error \eqref{eq:error_prob_2stage} for the decision rule \eqref{eq:detection_p_e_FC_legitimate_assumption} by minimizing the false alarm and missed detection probabilities. Any sequence of $0$'s and $1$'s can occur for the  detected trust vector $\mathbf{\hat{t}}$, each yielding a different error probability, so the error probability must be calculated for each possible vector $\mathbf{\hat{t}}$, along with each possible vector $\mathbf{y}$. Unfortunately, this computation scales exponentially with the number of robots, $N$. Furthermore, the true trust vector $\mathbf{t}$ and the probabilities of false alarm and missed detection of the malicious robots are unknown, i.e., $P_{\text{FA,M}}$ and $P_{\text{MD,M}}$, therefore, they cannot be used in minimizing  \eqref{eq:error_prob_2stage}.

To this end, we derive analytical guarantees regarding the error probability of the overall detection performance of the two-stage approach as follows. We find the worst-case probability of error of the FC by considering all the possible trust vectors $\mathbf{t}\in\{0,1\}^N$ and false alarm and missed detection probabilities $P_{\text{FA,M}}$ and $P_{\text{MD,M}}$, respectively, in the interval $[0,1]$, and choosing the $\mathbf{t}$, $P_{\text{FA,M}}$, and $P_{\text{MD,M}}$ that maximize \eqref{eq:error_prob_2stage}. Then, we minimize this worst-case error probability by choosing the best threshold $\gamma_t$, i.e., choose $\gamma_t = \gamma_t^*$ and tie-break probability $p_t = p_t^*$ where
\begin{equation}
    (\gamma_t^*,p_t^*) = \argmin_{\gamma_t,p_t} \max_{\mathbf{t},P_{\text{FA,M}},P_{\text{MD,M}}} P_{\text{e}}(\gamma_t,p_t,\mathbf{t},P_{\text{FA,M}},P_{\text{MD,M}}).
    \label{eq:opt_wors_case_two_stage}
\end{equation}

To this end, we must first determine the $P_{\text{FA,M}}, P_{\text{MD,M}}, \mathbf{t}$ that maximize $P_{\text{e}}$. In the remainder of this section, we assume that the proportion of malicious robots to expect in the network, denoted by $m$, is known, or we choose an upper bound for it $(\overline{m})$.

\begin{lemma} \label{lem:P_FA_M}
If $P_{\text{\emph{FA,L}}} < 0.5$ and $P_{\text{\emph{MD,L}}} < 0.5$, then the probability of false alarm and missed detection of the FC \eqref{eq:fa_md_total} is maximized for the Two Stage Approach when malicious robots choose $P_{\text{FA,M}} = P_{\text{MD,M}} = 1$, for any vector $\boldsymbol{t}\in\{0,1\}^N$.
\end{lemma}
\begin{proof}
 Recall the false alarm and missed detection probabilities for the FC using decision rules \eqref{eq:classification_decision} and \eqref{eq:detection_p_e_FC_legitimate_assumption} that lead to the overall false alarm and missed detection probabilities stated in \eqref{eq:fa_md_total}.

Next, we show that the false alarm probability \eqref{eq:fa_md_total} is maximized when $P_{\text{FA,M}} = 1$. The proof for $P_{\text{MD,M}}$ is analogous. In order to maximize $P_{\text{FA}}$ in \eqref{eq:fa_md_total} the summation must be maximized. We rewrite the summation by separating it into the terms affected by legitimate robots that were trusted and those affected by malicious robots that were trusted
\begin{equation}
\begin{aligned}
    &\sum_{i : \{ \hat{t}_i=1, t_i=0 \}}[w_{1,\text{L}} y_j - w_{0,\text{L}}(1-y_j)] + \\ &\sum_{i : \{ \hat{t}_i=1, t_i=1 \}} [w_{1,\text{L}} y_i - w_{0,\text{L}}(1-y_i)].
    \label{eq:lem_proof_P_FA_M_summation}
\end{aligned}
\end{equation}
Any robot $j \in \{ \hat{\mathcal{L}}\cap\mathcal{M} \}$ can maximize \eqref{eq:lem_proof_P_FA_M_summation} by maximizing $[w_{1,\text{L}} y_j - w_{0,\text{L}}(1-y_j)]$. Note that when $P_{\text{FA,L}} < 0.5$ and $P_{\text{MD,L}} < 0.5$ then $w_{1,\text{L}} > 0$ and $w_{0,\text{L}} > 0$. Thus, $[w_{1,\text{L}} y_j - w_{0,\text{L}}(1-y_j)]$ is maximized when $y_j = 1$ since $Y_j \in \{0,1\}$. Given the true hypothesis is $\mathcal{H}_0$, the measurement $Y_j = 1$ occurs when robot $j$ reports a false alarm. Therefore, the probability that robot $j$ reports $Y_j = 1$ is maximized when the probability of false alarm is maximized:
\begin{equation}
    \Pr(Y_j = 1 | \mathcal{H}_0) = P_{\text{FA,M}} = 1.
\end{equation}
\end{proof}

\begin{lemma} \label{lem:t_max}
Let $\overline{\mathbf{t}}$ be the worst-case vector $\mathbf{t}$, i.e., the vector $\mathbf{t}$ that maximizes the probability of error \eqref{eq:error_prob_2stage}. If $P_{\text{\emph{FA,L}}} < 0.5$, $P_{\text{\emph{MD,L}}} < 0.5$, and $P_{\text{\emph{FA,M}}} = P_{\text{\emph{MD,M}}} = 1$, then the probability of error $P_{\text{e}}(\gamma_t,p_t,\overline{\mathbf{t}},1,1)$ is maximized when $\overline{\mathbf{t}}$ contains the maximum number of malicious robots, i.e., \mbox{$\sum_{i\in\mathcal{N}}\overline{t}_i=N-\overline{m}N$.}
\end{lemma}
\begin{proof}
By Lemma~\ref{lem:P_FA_M} the probability of false alarm and missed detection \eqref{eq:fa_md_total} are maximized when a robot is trusted and its measurement reports the wrong hypothesis, i.e., $y_i = 1$ when the true event is $\Xi=0$ or $y_i = 0$ when the true event is $\Xi=1$. Since the optimal policy for malicious robots is to report the wrong hypothesis with probability $1$ (Lemma~\ref{lem:P_FA_M}), any robot increases the false alarm and missed detection probability of the FC when it is malicious instead of legitimate. Thus, the probability of error $P_{\text{e}}(\gamma_t,p_t,\mathbf{t},1,1)$ is maximized when the proportion of malicious robots, $m$, is maximized, i.e., when $\overline{\mathbf{t}}$ has $\overline{m}N$ malicious robots, where $\overline{m}$ is the upper bound on the proportion of malicious robots in the network.
\end{proof}

Utilizing Lemma~\ref{lem:t_max}, we calculate the exact probability of error for the FC for the worst-case attack where there are $\overline{m}N$ malicious robots and $P_{\text{FA,M}} = P_{\text{MD,M}} = 1$. In order to compute the probability of error exactly, we must compute the probability of false alarm and missed detection using \eqref{eq:fa_md_total}.
Let $k_{\text{L}} \in K_{\text{L}}$ be the number of legitimate robots trusted by the FC, where $K_{\text{L}} = \{ 0, \dots, (1-\overline{m})N \}$. Similarly, let $k_{\text{M}} \in K_{\text{M}}$ be the number of malicious robots trusted by the FC, where $K_{\text{M}} = \{ 0, \dots, \overline{m}N \}$. Let $S_N$ represent the left side of the inequalities in \eqref{eq:fa_md_total} given by:
\[S_\text{N} = \sum_{i=1}^N \hat{t}_i [w_{1,\text{L}} y_i - w_{0,\text{L}}(1-y_i)].\] 
Using the law of total probability, the false alarm  probability at the FC is given by
\begin{equation}
\begin{aligned}
    \resizebox{0.26\hsize}{!}{$P_{\text{FA}}(\gamma_t,p_t,\overline{m},1)$} &= \resizebox{0.68\hsize}{!}{$\sum_{k_{\text{L}} \in K_{\text{L}}, k_{\text{M}} \in K_{\text{M}}} \Pr(K_{\text{L}} = k_{\text{L}}) \Pr(K_{\text{M}} = k_{\text{M}})$} \\ & \qquad \qquad \quad \quad \quad \cdot \resizebox{0.42\hsize}{!}{$P_{\text{FA}}(S_\text{N} \geq \gamma_{\text{TS}} | \mathcal{H}_0, k_{\text{L}},k_{\text{M}})$}.
\end{aligned}
\end{equation}
Similarly, the probability of missed detection of the FC is given by
\begin{equation}
\begin{aligned}
    \resizebox{0.26\hsize}{!}{$P_{\text{MD}}(\gamma_t,p_t,\overline{m},1)$} &= \resizebox{0.68\hsize}{!}{$\sum_{k_{\text{L}} \in K_{\text{L}}, k_{\text{M}} \in K_{\text{M}}} \Pr(K_{\text{L}} = k_{\text{L}}) \Pr(K_{\text{M}} = k_{\text{M}})$} \\ & \qquad \qquad \quad \quad \quad \cdot \resizebox{0.42\hsize}{!}{$P_{\text{MD}}(S_\text{N} < \gamma_{\text{TS}} | \mathcal{H}_1, k_{\text{L}},k_{\text{M}})$}.
\end{aligned}
\end{equation}
The probability of false alarm for a particular instantiation of $k_{\text{L}}$ and $k_{\text{M}}$ can be written as a function of the binomial Cumulative Distribution Function:
\begin{equation}
    \begin{aligned}
        &P_{\text{FA}}(S_\text{N} \geq \gamma_{\text{TS}} | \mathcal{H}_0, k_{\text{L}},k_{\text{M}}) \\ &= \resizebox{0.87\hsize}{!}{$\Pr\left( \sum_{i: \{\hat{t}_i=1,t_i=1\}} y_i \geq \frac{ \gamma_{\text{TS}} - k_{\text{M}}w_{1,\text{L}} + k_{\text{L}}w_{0,\text{L}} }{w_{0,\text{L}} + w_{1,\text{L}}} \Big| \mathcal{H}_0,k_{\text{L}},k_{\text{M}}, \right)$}, \\ &= 1 - F_{\text{b}}\left( \lceil\frac{ \gamma_{\text{TS}} - k_{\text{M}}w_{1,\text{L}} + k_{\text{L}}w_{0,\text{L}} }{w_{0,\text{L}} + w_{1,\text{L}}}\rceil - 1 ; P_{\text{FA,L}},k_{\text{L}}\right),
    \end{aligned}
    \label{eq:P_FA_kl_km}
\end{equation}
where $F_{\text{b}}(g;p,n) = \sum_{i=0}^g \binom{n}{i} p^i (1-p)^{n-i} = \Pr\left( \sum_{i=1}^n y_i \leq g \right)$ is the binomial Cumulative Distribution Function evaluated at $g$ for $n$ variables and success probability $p$. Similarly, for the probability of missed detection, we have that
\begin{equation}
\begin{aligned}
    &P_{\text{MD}}(S_\text{N} < \gamma_{\text{TS}} | \mathcal{H}_1, k_{\text{L}},k_{\text{M}})= \\ &   F_{\text{b}}\left( \lceil\frac{ \gamma_{\text{TS}} + k_{\text{M}}w_{0,\text{L}} + k_{\text{L}}w_{0,\text{L}} }{w_{0,\text{L}} + w_{1,\text{L}}}\rceil - 1 ; 1-P_{\text{MD,L}},k_{\text{L}}\right).
\end{aligned}
\end{equation}
 
Recall \eqref{eq:trust_prob_two_stage}. We note that these probabilities depend on the distribution of the robot's vector of trust values $\boldsymbol{a}$. Then, we have that
\begin{equation}
    \begin{aligned}
        \Pr(K_{\text{L}} = k_{\text{L}}) &= \Pr\left( \sum_{i \in \mathcal{L}} \hat{t}_i = k_{\text{L}} \right) \\ &= f_{\text{b}}(k_{\text{L}} ; P_{\text{trust,L}}(\gamma_t,p_t), (1-\overline{m})N), \\ \Pr(K_{\text{M}} = k_{\text{M}}) &= \Pr\left( \sum_{i \in \overline{\mathcal{M}}} \hat{t}_i = k_{\text{M}} \right) \\ &= f_{\text{b}}(k_{\text{M}} ; P_{\text{trust,M}}(\gamma_t,p_t), \overline{m}N),
    \end{aligned}
\end{equation}
where $f_{\text{b}}(g ; p,n) = \binom{n}{g} p^g (1-p)^{n-g} = \Pr\left( \sum_{i=1}^n y_i = g \right)$ is the binomial probability distribution function evaluated at $g$ for $n$ variables and success probability $p$. Thus, the probability of false alarm and missed detection are
\begin{equation}
    \begin{aligned}
        &P_{\text{FA}}(\gamma_t,p_t,\overline{m},1) \\
        &= \sum_{k_{\text{L}} \in K_{\text{L}}, k_{\text{M}} \in K_{\text{M}}}  f_{\text{b}}(k_{\text{L}} ; P_{\text{trust,L}}(\gamma_t,p_t), (1-\overline{m})N) \cdot \\ 
        &\hspace{2cm} f_{\text{b}}(k_{\text{M}} ; P_{\text{trust,M}}(\gamma_t,p_t), \overline{m}N) \cdot \\ 
        &\hspace{2cm} P_{\text{FA}}(S_\text{N} \geq \gamma_{\text{TS}} | \mathcal{H}_0, k_{\text{L}},k_{\text{M}}), \\ 
        &P_{\text{MD}}(\gamma_t,p_t,\overline{m},1) \\
        &= \sum_{k_{\text{L}} \in K_{\text{L}}, k_{\text{M}} \in K_{\text{M}}}  f_{\text{b}}(k_{\text{L}} ; P_{\text{trust,L}}(\gamma_t,p_t), (1-\overline{m})N) \cdot \\ 
        &\hspace{2cm} f_{\text{b}}(k_{\text{M}} ; P_{\text{trust,M}}(\gamma_t,p_t), \overline{m}N) \cdot \\ 
        &\hspace{2cm} P_{\text{MD}}(S_\text{N} < \gamma_{\text{TS}} | \mathcal{H}_1, k_{\text{L}},k_{\text{M}}).
    \end{aligned}
    \label{eq:P_FA_MD_WC}
\end{equation}

Therefore, we define the total error probability in the worst-case
\begin{equation}
\begin{aligned}
    \overline{P}_{\text{e}}(\gamma_t,p_t,\overline{m},1,1) \triangleq &\Pr(\Xi=0) P_{\text{FA}}(\gamma_t,p_t,\overline{m},P_{\text{FA,M}}=1) + \\ & \Pr(\Xi=1) P_{\text{MD}}(\gamma_t,p_t,\overline{m},P_{\text{MD,M}}=1),
\end{aligned}
\label{eq:P_e_2SA_alg}
\end{equation}
and we can choose the thresholds $\gamma_t$ and $p_t$ that minimize the expression. Once we choose the thresholds $\gamma_t$ and $p_t$, the rest of the Two Stage Approach becomes a standard binary hypothesis testing problem.

\begin{lemma}\label{lemma:sufficient_discrete_gamma}
Denote \[\Gamma_t: = \left\{ \frac{p_{\alpha}(a|t_i=1)}{p_{\alpha}(a|t_i=0)} \right\}_{a\in\mathcal{A}},\]
where $\{\cdot\}_{a\in\mathcal{A}}$ represents a set consisting of all possible values of $a\in\mathcal{A}$ and the set $\mathcal{A}$ follows \cref{assumption:trust_val}. Then, the minimal value of \eqref{eq:opt_wors_case_two_stage} with respect to $\gamma_t$ can be achieved by $\gamma_t\in\Gamma_t$.
\end{lemma}
\begin{proof}
The proof follows directly from the finiteness of the set $\mathcal{A}$ and since $p_t$ can take values in the interval $[0,1]$.
\end{proof}

\begin{algorithm}[h!]
\caption{Two Stage Approach \\ Input: $P_{\text{FA},\text{L}}$, $P_{\text{MD},\text{L}}$, $\hat{P}_{\text{FA},\text{M}} = \hat{P}_{\text{MD},\text{M}} = 1$, $\Pr(\Xi=0)$, $\Pr(\Xi=1)$, $\mathbf{y}$, $\boldsymbol{a}$, $\overline{m}$, $\Gamma_t$, $\delta_p$ \\ Output: Decision $\mathcal{H}_0$ or $\mathcal{H}_1$}
\label{alg:2stage_discrete}
\begin{algorithmic}[1]
\State Set $\Gamma_p = \{0,\delta_p,2\delta_p,\dots,1\}$.
\State Set $\gamma_{t,\text{temp}}=0$, $p_{t,\text{temp}}=0$, $P_{\text{e,temp}}=2$.
\ForAll{$\hat{\gamma}_t \in \Gamma_t$, $\hat{p}_t \in \Gamma_p$}
\State Compute $P_{\text{trust,L}}(\hat{\gamma}_t,\hat{p}_t)$, $P_{\text{trust,M}}(\hat{\gamma}_t,\hat{p}_t)$ by \eqref{eq:trust_prob_two_stage}.
\State Compute $P_{\text{FA}}(\hat{\gamma}_t,\hat{p}_t,\overline{m},1)$, $P_{\text{MD}}(\hat{\gamma}_t,\hat{p}_t,\overline{m},1)$ by \eqref{eq:P_FA_MD_WC}.
\State Compute $\overline{P}_{\text{e}}(\hat{\gamma}_t,\hat{p}_t,\overline{m},1,1)$ by \eqref{eq:P_e_2SA_alg}. 
\If{$\overline{P}_{\text{e}}(\hat{\gamma}_t,\hat{p}_t,\overline{m},1,1)<P_{\text{e,temp}}$}
\State Set $(\gamma_{t,\text{temp}}, p_{t,\text{temp}})=(\hat{\gamma}_t,\hat{p}_t)$.
\State Set $P_{\text{e,temp}}=\overline{P}_{\text{e}}(\hat{\gamma}_t,\hat{p}_t,\overline{m},1,1)$.
\EndIf
\EndFor
\State Set $(\gamma_t, p_t) = (\gamma_{t,\text{temp}}, p_{t,\text{temp}})$.
\State Determine the vector $\mathbf{\hat{t}}$ using \eqref{eq:classification_decision}.
\State Determine decision using \eqref{eq:detection_p_e_FC_legitimate_assumption}.
\State Return decision $\mathcal{H}_0$ or $\mathcal{H}_1$.
\end{algorithmic}
\end{algorithm}

Let $\Gamma_p := \{0,\delta_p,2\delta_p,\dots,1\}$ with a discretization constant $\delta_p$. Algorithm~\ref{alg:2stage_discrete} explains the Two Stage Approach step-by-step. Algorithm~\ref{alg:2stage_discrete} takes a set $\Gamma_t$ as input. Then, for each $\hat{\gamma}_t \in \Gamma_t$ and each $\hat{p}_t \in \Gamma_p$ we compute $P_{\text{trust,L}}(\hat{\gamma}_t,\hat{p}_t)$, $P_{\text{trust,M}}(\hat{\gamma}_t,\hat{p}_t)$, as well as $P_{\text{FA}}(\hat{\gamma}_t,\hat{p}_t,\overline{m},1)$ and $P_{\text{MD}}(\hat{\gamma}_t,\hat{p}_t,\overline{m},1)$. Then we compute the probability of error at the FC for the given $\hat{\gamma}_t$ and $\hat{p}_t$. The $\hat{\gamma}_t$ and $\hat{p}_t$ that yields the minimum probability of error is then used in the decision rule in \eqref{eq:classification_decision} to determine which robots to trust or not trust (vector $\mathbf{\hat{t}}$). Finally, we use the chosen vector $\mathbf{\hat{t}}$ to make a decision using the standard binary hypothesis decision rule \eqref{eq:detection_p_e_FC_legitimate_assumption}.

Determining the threshold value $\gamma_t$ and tie-break probability $p_t$ requires computing the probability of error $|\Gamma_t|\cdot|\Gamma_p|$ times, where $|\cdot|$ represents the cardinality of the set. However, this only needs to be computed once, and then the returned $\gamma_t$ and $p_t$ can be used to run each subsequent hypothesis test. With a given $\gamma_t$ and $p_t$, the hypothesis test requires $\mathcal{O}(N)$ comparisons.

\begin{theorem}\label{thm:2SA_optimum_decision_rule}
Assume that the FC uses the decision rule in \eqref{eq:classification_decision} to detect malicious robots, and then uses the decision rule \eqref{eq:detection_p_e_FC_legitimate_assumption}. Then Algorithm~\ref{alg:2stage_discrete} chooses the threshold value $\gamma_t$ and tie-break probability $p_t$ that minimize the worst-case probability of error of the FC up to a  discretization distance \[d(\delta_p):=\min_{p_t\in\Gamma_p}\overline{P}_{\text{e}}(\gamma_t^*,p_t,\overline{m},1,1)-\overline{P}_{\text{e}}(\gamma_t^*,p_t^*,\overline{m},1,1).\]
Furthermore, $d(\delta_p)\rightarrow 0$ as $\delta_p\rightarrow 0$.
\end{theorem}
\begin{proof}
The goal is to minimize the worst-case probability of error of the FC, i.e.,
\begin{equation}
    \min_{\gamma_t,p_t} \max_{\mathbf{t},P_{\text{FA,M}},P_{\text{MD,M}}} P_{\text{e}}(\gamma_t,p_t,\mathbf{t},P_{\text{FA,M}},P_{\text{MD,M}}).
\end{equation}
Using the results from Lemmas~\ref{lem:P_FA_M},~\ref{lem:t_max} and \eqref{eq:P_FA_MD_WC} we upper bound the error probability using the worst-case error probability:
\begin{equation}
    \begin{aligned}
        \min_{\gamma_t,p_t} \max_{\mathbf{t},P_{\text{FA,M}},P_{\text{MD,M}}} P_{\text{e}}&(\gamma_t,p_t,\mathbf{t},P_{\text{FA,M}},P_{\text{MD,M}}) \\ &= \min_{\gamma_t,p_t} \max_{\mathbf{t}} P_{\text{e}}(\gamma_t,p_t,\mathbf{t},1,1), \\ 
        &= \min_{\gamma_t,p_t} \overline{P}_e(\gamma_t,p_t,\overline{m},1,1).
    \end{aligned}
\end{equation}
The equality in the first line directly follows from Lemma~\ref{lem:P_FA_M}. The second line follows from the first by inserting the worst-case vector $\mathbf{t}$, with $\overline{m}$ malicious robots, as the one that maximizes the probability of error $P_{\text{e}}$ (Lemma~\ref{lem:t_max}).

Additionally, by Lemma~\ref{lemma:sufficient_discrete_gamma}, it is  sufficient to optimize $\gamma_t$ over the set  $\Gamma_t$. 
Now, since we optimize $p_t$ using a line search, we may not necessarily find an optimal pair $(\gamma_t^*,p_t^*)$. However, we can express the distance from the optimal solution for the worst case scenario by:
\begin{flalign}
    &\min_{\gamma_t\in\Gamma_t,p_t\in\Gamma_p}\overline{P}_{\text{e}}(\gamma_t,p_t,\overline{m},1,1)-\overline{P}_{\text{e}}(\gamma_t^*,p_t^*,\overline{m},1,1)\nonumber\\
    &\leq \min_{p_t\in\Gamma_p}\overline{P}_{\text{e}}(\gamma_t^*,p_t,\overline{m},1,1)-\overline{P}_{\text{e}}(\gamma_t^*,p_t^*,\overline{m},1,1)\nonumber\\
    &=d(\delta_p).
\end{flalign}
For every fixed $\gamma_t$, the function $\overline{P}_{\text{e}}(\gamma_t,p_t,\overline{m},1,1)$ is a polynomial function of $p_t$, therefore, it is continuous in $p_t$ (over the interval $p_t\in[0,1]$). 
Consequently, 
$d(\delta_p)\rightarrow 0$ as $\delta_p\rightarrow 0$. 
\end{proof}

\subsection{Error Bounds for the Two Stage Approach}

In this section, we show that when the probability of the FC trusting a legitimate robot in the first stage of the Two Stage Approach \eqref{eq:classification_decision} is much greater than the probability of the FC trusting a malicious robot, that the overall probability of error at the FC decreases towards $0$ as the number of robots in the network increases. To  this end, we derive an upper bound for the probability of error when using the Two Stage Approach \eqref{eq:P_e_2SA_alg} and show that the probability of error decays at least exponentially as the number of robots approaches $\infty$.

Let $\beta_\text{M} \in (0,1)$ and $\beta_\text{L} \in (0,1)$ denote the proportion of malicious (resp. legitimate) robots that are trusted by the FC after the first stage of the Two Stage Approach. The terms $\beta_\text{M}$ and $\beta_\text{L}$ are purely for analytical purposes. They will be utilized to split the probability of error analysis into four separate events, corresponding to differing numbers of trusted legitimate and malicious robots. 

Let us consider a given threshold value $\gamma_t$ and tie-break probability $p_t$ at the first stage, and let $P_{\text{trust,M}}(\gamma_t,p_t)$ and $P_{\text{trust,L}}(\gamma_t,p_t)$ be the resulting probability of trusting a malicious (resp. legitimate) robot.
Furthermore, let us consider a given $\beta_{\text{M}}$ and $\beta_{\text{L}}$ such that $\beta_\text{M} > P_{\text{trust,M}}(\gamma_t,p_t)$ and $\beta_\text{L} < P_{\text{trust,L}}(\gamma_t,p_t)$.  Intuitively, the values $P_{\text{trust,M}}(\gamma_t,p_t)$ and $P_{\text{trust,L}}(\gamma_t,p_t)$ correspond to the expected proportion of malicious and legitimate robots that will be trusted by the FC. Consequently, when we consider 
\[\beta_\text{M} > P_{\text{trust,M}}(\gamma_t,p_t) \text{ and } \beta_\text{L} < P_{\text{trust,L}}(\gamma_t,p_t)\] we are representing undesirable regions where more than the expected proportion of malicious robots are trusted and less than the expected proportion of legitimate robots are trusted. Finally, assume \[\beta_{\text{L}}|\mathcal{L}| >> \max\{ \beta_{\text{M}}|\mathcal{M}|,1 \}.\] 
This means we consider scenarios where many more legitimate robots than malicious robots are trusted by the FC. This is likely to occur when 
\[P_{\text{trust,L}}(\gamma_t,p_t) >> P_{\text{trust,M}}(\gamma_t,p_t).\]

Recall that $k_{\text{L}}$ and $k_{\text{M}}$ denote the actual number of legitimate and malicious robots trusted by the FC. In what follows, we upper bound the probability of error by examining four distinct cases, each considering a different regime with respect to the number of trusted legitimate and malicious robots:
\begin{enumerate}
    \item $k_{\text{L}} \leq \beta_{\text{L}}|\mathcal{L}|$, $k_{\text{M}} < \beta_{\text{M}}|\mathcal{M}|$,
    \item $k_{\text{L}} \leq \beta_{\text{L}}|\mathcal{L}|$, $k_{\text{M}} \geq \beta_{\text{M}}|\mathcal{M}|$,
    \item $k_{\text{L}} > \beta_{\text{L}}|\mathcal{L}|$, $k_{\text{M}} \geq \beta_{\text{M}}|\mathcal{M}|$,
    \item $k_{\text{L}} > \beta_{\text{L}}|\mathcal{L}|$, $k_{\text{M}} < \beta_{\text{M}}|\mathcal{M}|$.
\end{enumerate} 
In words, these cases correspond to scenarios where 1) few legitimate and malicious robots are trusted after the first stage of the Two Stage Approach, 2) few legitimate robots are trusted but many malicious robots are trusted, 3) many legitimate and malicious robots are trusted, and 4) many legitimate robots are trusted but few malicious robots are trusted. Intuitively, Cases 1, 2, and 3 will contribute the most to the detection error probability since they contain either many malicious robots or few legitimate robots, whereas the fourth event is the most desirable since it contains many legitimate robots and few malicious robots. In what follows, we investigate scenarios where the probabilities corresponding to Cases 1, 2, or 3 occurring decay at least exponentially as the number of robots increases, then show that the probability of error with respect to Case 4 also decays at least exponentially as the number of robots increases.

Recall that $\overline{m}$ is the upper bound on the true proportion of malicious robots in the network, $\gamma_{\text{TS}} = \log{\left( \frac{\Pr(\Xi=0)}{\Pr(\Xi=1)}\right)}$ is the decision threshold used in the second stage of the Two Stage Approach \eqref{eq:detection_p_e_FC_legitimate_assumption}, $\hat{t}_i$ denotes the outcome of the first stage which tests the trustworthiness of robot $i$, and
\begin{equation*}
    S_\text{N} = \sum_{i=1}^N \hat{t}_i [w_{1,\text{L}} y_i - w_{0,\text{L}}(1-y_i)],
\end{equation*}
is the left side of the inequalities in \eqref{eq:fa_md_total}. Assume $\overline{m} \in (0,1)$. The probability of a particular case occurring corresponds to the probability of trusting $k_{\text{L}}$ and $k_{\text{M}}$ robots that fall into the region described by the particular case. 

These probabilities are conditioned upon the chosen threshold values $\gamma_t$ and $p_t$, but we omit these threshold values from the case probabilities for ease of exposition. With these four cases, we can upper bound the worst-case probability of error by using the union bound as follows:
    \begin{flalign}
        &\overline{P}_\text{e}(\gamma_t,p_t,\overline{m},P_{\text{FA,M}}=1,P_{\text{MD,M}}=1)\nonumber\\
        &\leq \Pr(k_{\text{L}} \leq \beta_{\text{L}}|\mathcal{L}|) \Pr(k_{\text{M}} < \beta_{\text{M}}|\mathcal{M}|) \max_{\substack{k_{\text{L}}\leq \beta_{\text{L}}|\mathcal{L}|,\\ k_{\text{M}}< \beta_{\text{M}}|\mathcal{M}|}}\overline{p}_{\text{e}}(k_{\text{L}}, k_{\text{M}})  \nonumber\\ 
        &+ \Pr(k_{\text{L}} \leq \beta_{\text{L}}|\mathcal{L}|) \Pr(k_{\text{M}} \geq \beta_{\text{M}}|\mathcal{M}|)) \max_{\substack{k_{\text{L}}\leq \beta_{\text{L}}|\mathcal{L}|,\\ k_{\text{M}}\geq \beta_{\text{M}}|\mathcal{M}|}}\overline{p}_{\text{e}}(k_{\text{L}}, k_{\text{M}}) \nonumber\\ 
        &+ \Pr(k_{\text{L}} > \beta_{\text{L}}|\mathcal{L}|) \Pr(k_{\text{M}} \geq \beta_{\text{M}}|\mathcal{M}|)  \max_{\substack{k_{\text{L}}> \beta_{\text{L}}|\mathcal{L}|,\\ k_{\text{M}}\geq \beta_{\text{M}}|\mathcal{M}|}}\overline{p}_{\text{e}}(k_{\text{L}}, k_{\text{M}}) \nonumber\\
        &+ \Pr(k_{\text{L}} > \beta_{\text{L}}|\mathcal{L}|) \Pr(k_{\text{M}} < \beta_{\text{M}}|\mathcal{M}|)  \max_{\substack{k_{\text{L}}> \beta_{\text{L}}|\mathcal{L}|,\\ k_{\text{M}}< \beta_{\text{M}}|\mathcal{M}|}}\overline{p}_{\text{e}}(k_{\text{L}}, k_{\text{M}}), \label{eq:long_UB1_unsimplified}
\end{flalign}
where 
\begin{flalign}
     \overline{p}_{\text{e}}(k_{\text{L}}, k_{\text{M}})&\triangleq \Pr(\Xi=0) P_{\text{FA}}(S_\text{N} \geq \gamma_{\text{TS}} | \mathcal{H}_0, k_{\text{L}},k_{\text{M}}) \nonumber\\
     &+  \Pr(\Xi=1) P_{\text{MD}}(S_\text{N} < \gamma_{\text{TS}} | \mathcal{H}_1, k_{\text{L}},k_{\text{M}}),
     \end{flalign} 
represents the probability of error for a given $k_{\text{L}}$ and $k_{\text{M}}$ corresponding to a particular case. Note that $\overline{p}_{\text{e}}(k_{\text{L}}, k_{\text{M}})\leq 1$ for any $k_{\text{L}}$ and $k_{\text{M}}$. Consequently, we can simplify \eqref{eq:long_UB1_unsimplified} to
    \begin{flalign}
        &\overline{P}_\text{e}(\gamma_t,p_t,\overline{m},1,1)\nonumber\\
        &\leq  \Big[ \Pr(k_{\text{L}} \leq \beta_{\text{L}}|\mathcal{L}|) \Pr(k_{\text{M}} < \beta_{\text{M}}|\mathcal{M}|)  \nonumber\\ 
        &+ \Pr(k_{\text{L}} \leq \beta_{\text{L}}|\mathcal{L}|) \Pr(k_{\text{M}} \geq \beta_{\text{M}}|\mathcal{M}|)) \Big]\cdot 1 \nonumber\\ 
        &+ \Pr(k_{\text{L}} > \beta_{\text{L}}|\mathcal{L}|) \Pr(k_{\text{M}} \geq \beta_{\text{M}}|\mathcal{M}|)\cdot 1 \nonumber\\
        &+ \Pr(k_{\text{L}} > \beta_{\text{L}}|\mathcal{L}|) \Pr(k_{\text{M}} < \beta_{\text{M}}|\mathcal{M}|)  \max_{\substack{k_{\text{L}}> \beta_{\text{L}}|\mathcal{L}|,\\ k_{\text{M}}< \beta_{\text{M}}|\mathcal{M}|}}\overline{p}_{\text{e}}(k_{\text{L}}, k_{\text{M}}).
        \label{eq:long_UB1}
    \end{flalign}
We utilize the upper bound $\overline{p}_{\text{e}}(k_{\text{L}},k_{\text{M}})\leq 1$ for the cases where few legitimate robots are trusted, $k_{\text{L}}\leq \beta_{\text{L}}|\mathcal{L}|$, or many malicious robots are trusted, $k_{\text{M}} \geq \beta_{\text{M}}|\mathcal{M}|$, to simplify the analysis since these represent cases where the probability of error is likely high. We intend to show that the probability of these cases occurring decays at least exponentially as the number of robots increases.

Let
\begin{equation}
\begin{aligned}
    \Pr(\text{Case 1}) &\triangleq \Pr(k_{\text{L}} \leq \beta_{\text{L}}|\mathcal{L}|) \Pr(k_{\text{M}} < \beta_{\text{M}}|\mathcal{M}|), \\ \Pr(\text{Case 2}) &\triangleq \Pr(k_{\text{L}} \leq \beta_{\text{L}}|\mathcal{L}|) \Pr(k_{\text{M}} \geq \beta_{\text{M}}|\mathcal{M}|), \\
    \Pr(\text{Case 3}) &\triangleq \Pr(k_{\text{L}} > \beta_{\text{L}}|\mathcal{L}|) \Pr(k_{\text{M}} \geq \beta_{\text{M}}|\mathcal{M}|), \\ \Pr(\text{Case 4}) &\triangleq \Pr(k_{\text{L}} > \beta_{\text{L}}|\mathcal{L}|) \Pr(k_{\text{M}} < \beta_{\text{M}}|\mathcal{M}|),
\end{aligned}
\label{eq:cases_def}
\end{equation}
be the probability of Case 1, Case 2, Case 3, and Case 4 occurring, respectively. We are interested in how the probability of error in \eqref{eq:long_UB1} is affected when the number of robots increases. To see this more clearly, we rewrite \eqref{eq:long_UB1} using \eqref{eq:cases_def}, and then analyze each term separately:
\begin{equation}
\begin{aligned}
    &\overline{P}_\text{e}(\gamma_t,p_t,\overline{m},1,1) \leq \left[ \Pr(\text{Case 1}) + \Pr(\text{Case 2}) \right] \\ &+ \Pr(\text{Case 3}) + \Pr(\text{Case 4}) \cdot \max_{k_{\text{L}}> \beta_{\text{L}}|\mathcal{L}|, k_{\text{M}}< \beta_{\text{M}}|\mathcal{M}|}\overline{p}_{\text{e}}(k_{\text{L}}, k_{\text{M}}).
\end{aligned}
\label{eq:UB_intuitive}
\end{equation}

We derive our upper bound \eqref{eq:UB_intuitive} on the error probability by examining the terms within. To this end, we utilize the following upper bound \cite{Chernoff} which is derived from the Chernoff bound
\begin{equation}\label{eq:chernoff_upper}
    \Pr(X \leq g) = F_{\text{b}}(g;n,p) \leq \exp\left(-n D\left(\frac{g}{n}||p\right)\right),
\end{equation}
where we assume $X$ to be a binomial distribution, $n$ is the number of trials, $p$ is the success probability, i.e., the probability a trial results in a $1$, and 
\[D(p||q) = p \log\left(\frac{p}{q}\right) + (1-p) \log\left(\frac{1-p}{1-q}\right)\] 
is the Kullback–Leibler (KL) divergence between a Bernoulli random variable with success probability $p$ and a Bernoulli random variable with success probability $q$. The Chernoff bound \eqref{eq:chernoff_upper} provides an upper bound for the lower tail of the cumulative distribution function for $\Pr(X \leq g)$, and is valid when $\frac{g}{n} \in (0,p)$. The Chernoff bound can also provide an upper bound for the upper tail of the cumulative distribution function for $\Pr(X \geq g)$ for $\frac{g}{n} \in (p,1)$.

Next, we analyze the terms within \eqref{eq:UB_intuitive}. Specifically, for Cases 1, 2, and 3 we show that the probability of them occurring decays at least exponentially as the number of robots increases. Then, we show that the probability of Case 4 occurring approaches $1$ as the number of robots increases, but that the corresponding probability of error for Case 4 decays at least exponentially.

\paragraph{\textbf{Cases 1 and 2}} Cases 1 and 2 correspond to cases where few legitimate robots are trusted by the FC. We show that the probability of Cases 1 or 2 occurring decays at least exponentially as the number of robots increases. First, we simplify the probability of Cases 1 or 2 occurring using the law of total probability:
\begin{equation*}
\begin{aligned}
    \Pr(\text{Case 1}&) + \Pr(\text{Case 2}) \\ &= \Pr(k_{\text{L}} \leq \beta_{\text{L}}|\mathcal{L}|) \Pr(k_{\text{M}} < \beta_{\text{M}}|\mathcal{M}|) \\ 
        &\quad + \Pr(k_{\text{L}} \leq \beta_{\text{L}}|\mathcal{L}|) \Pr(k_{\text{M}} \geq \beta_{\text{M}}|\mathcal{M}|)) \\ &= \Pr(k_{\text{L}} \leq \beta_{\text{L}}|\mathcal{L}|).
\end{aligned}
\end{equation*}

Next, observe that the number of trusted legitimate robots, i.e., $k_{\text{L}} = \sum_{i \in \mathcal{L}} \hat{t}_i$, is distributed according to a binomial distribution with the probability for $\hat{t}_i = 1$ equal to the probability of trusting a legitimate robot $i \in \mathcal{L}$. Therefore, the upper bound on $\Pr(k_\text{L} \leq \beta_\text{L}|\mathcal{L}|)$ can be written by
\begin{equation}
\begin{aligned}
        \Pr(k_\text{L} \leq \beta_\text{L} |\mathcal{L}|) &\leq \exp\left(-|\mathcal{L}| D\left( \beta_\text{L} || P_{\text{trust,L}}(\gamma_t,p_t) \right)\right) \\ &\leq \exp\left(-(1-\overline{m})N D\left( \beta_\text{L} || P_{\text{trust,L}}(\gamma_t,p_t) \right)\right).
\end{aligned}
\label{eq:loss_legit_bounds}
\end{equation}
The Chernoff bound is valid here since we consider the region where $\beta_\text{L} < P_{\text{trust,L}}(\gamma_t,p_t)$. It can be seen from \eqref{eq:loss_legit_bounds} that the upper bound on the probability of Case 1 or Case 2 occurring decays exponentially with a rate of $(1-\overline{m})ND\left( \beta_\text{L} || P_{\text{trust,L}}(\gamma_t,p_t) \right)$ assuming $\overline{m} \neq 1$. This is guaranteed to be an exponential decay because the KL divergence is always non-negative, $\beta_\text{L} \neq P_{\text{trust,L}}(\gamma_t,p_t)$, and $N > 0$.

\paragraph{\textbf{Case 3}} Case 3 corresponds to the case where many legitimate robots are trusted by the FC, but also many malicious robots are trusted. Similar to Cases 1 and 2, we show that the probability of Case 3 occurring decays at least exponentially as the number of robots increases. Recall that
\begin{equation*}
    \Pr(\text{Case 3}) = \Pr(k_{\text{L}} > \beta_{\text{L}}|\mathcal{L}|) \Pr(k_{\text{M}} \geq \beta_{\text{M}}|\mathcal{M}|).
\end{equation*}
For Cases 1 and 2 we showed in \eqref{eq:loss_legit_bounds} that $\Pr(k_{\text{L}} \leq \beta_{\text{L}}|\mathcal{L}|)$ decays toward 0 at least exponentially as the number of robots increases. 
Since
\begin{equation*}
    \Pr(k_{\text{L}} > \beta_{\text{L}}|\mathcal{L}|) = 1 - \Pr(k_{\text{L}} \leq \beta_{\text{L}}|\mathcal{L}|)\leq 1,
\end{equation*}
we conclude by the sandwich theorem~\cite{clarke1994mean} that $\Pr(k_{\text{L}} > \beta_{\text{L}}|\mathcal{L}|)$ approaches $1$ as $N$ tends to infinity. However, observe that the number of trusted malicious robots, i.e., $k_{\text{M}} = \sum_{i \in \mathcal{M}} \hat{t}_i$, is distributed according to a binomial distribution with the probability for $\hat{t}_i = 1$ equal to the probability of trusting a robot $i$ given that $i \in \mathcal{M}$. Then, using the Chernoff bound \eqref{eq:chernoff_upper}, the upper bound on $\Pr(k_{\text{M}} \geq \beta_{\text{M}}|\mathcal{M}|)$ can be written by
\begin{flalign}
        \Pr(k_\text{M} \geq \beta_\text{M} |\mathcal{M}|) &\leq \exp\left(-|\mathcal{M}| D\left( \beta_\text{M} || P_{\text{trust,M}}(\gamma_t,p_t) \right)\right) \nonumber\\ &\leq \exp\left(-\overline{m}N D\left( \beta_\text{M} || P_{\text{trust,M}}(\gamma_t,p_t) \right)\right).
        \label{eq:loss_malicious_bounds}
\end{flalign}
The Chernoff bound is valid here since we consider the region where $\beta_\text{M} > P_{\text{trust,M}}(\gamma_t,p_t)$. It can be seen from \eqref{eq:loss_malicious_bounds} that the upper bound on $\Pr(k_\text{M} \geq \beta_\text{M} |\mathcal{M}|)$, and thus the probability of Case 3 occurring, decays exponentially with a rate of $\overline{m}ND\left( \beta_\text{M} || P_{\text{trust,M}}(\gamma_t,p_t) \right)$ assuming $\overline{m} \neq 0$. Again, this is guaranteed to be an exponential decay because the KL divergence is always non-negative, $\beta_\text{M} \neq P_{\text{trust,M}}(\gamma_t,p_t)$, and $N > 0$.

\paragraph{\textbf{Case 4}} Case 4 is the ideal case, where many legitimate robots are trusted by the FC and few malicious robots are trusted. We show that the probability of Case 4 occurring approaches $1$ as the number of robots increases, but the corresponding probability of error decays at least exponentially. Recall that
\begin{equation*}
    \Pr(\text{Case 4}) = \Pr(k_{\text{L}} > \beta_{\text{L}}|\mathcal{L}|) \Pr(k_{\text{M}} < \beta_{\text{M}}|\mathcal{M}|).
\end{equation*}
We already showed that $\Pr(k_{\text{L}} > \beta_{\text{L}}|\mathcal{L}|) \rightarrow 1$ as $N \rightarrow \infty$. Similarly, $\Pr(k_{\text{M}} < \beta_{\text{M}}|\mathcal{M}|) \rightarrow 1$ as $N \rightarrow \infty$ since \mbox{$\Pr(k_{\text{M}} \geq \beta_{\text{M}}|\mathcal{M}|) \rightarrow 0$}.

For Case 4 we must also upper bound the probability of error, which requires upper bounding the probability of false alarm and missed detection for a given $k_{\text{L}}$ and $k_{\text{M}}$. For both of these, we use the Chernoff bound again. 

First, we analyze the false alarm probability.
Recall the form in \eqref{eq:P_FA_kl_km} which allows us to write the upper bound as
\begin{equation}
    \begin{aligned}
        P_{\text{FA}}&(S_\text{N} \geq \gamma_{\text{TS}} | \mathcal{H}_0, k_{\text{L}},k_{\text{M}}) \\ &= \resizebox{0.9\hsize}{!}{$\Pr\left( \sum_{i: \{\hat{t}_i=1,t_i=1\}} y_i \geq \frac{ \gamma_{\text{TS}} - k_{\text{M}}w_{1,\text{L}} + k_{\text{L}}w_{0,\text{L}} }{w_{0,\text{L}} + w_{1,\text{L}}} \Big| \mathcal{H}_0,k_{\text{L}},k_{\text{M}} \right)$} \\ & \leq \exp\left( -k_\text{L} D\left( \tilde{\gamma}_{\text{FA}}(k_{\text{L}},k_{\text{M}}) || P_{\text{FA,L}} \right) \right),
    \end{aligned}
    \label{eq:loss_FA_bounds}
\end{equation}
where
\begin{equation}
    \tilde{\gamma}_{\text{FA}}(k_{\text{L}},k_{\text{M}}) := \frac{1}{k_{\text{L}}} \frac{\gamma_{\text{TS}} - k_{\text{M}}w_{1,\text{L}} + k_{\text{L}}w_{0,\text{L}} }{(w_{0,\text{L}} + w_{1,\text{L}})},
    \label{eq:tilde_FA}
\end{equation}
is the threshold on the right-hand side of the inequality in \eqref{eq:P_FA_kl_km} and \eqref{eq:loss_FA_bounds} normalized with respect to the number of trusted legitimate robots $k_{\text{L}}$.

Similarly, the probability of missed detection given $k_\text{L}$ and $k_\text{M}$ is upper bounded by
\begin{equation}
\begin{aligned}
        P_{\text{MD}}(S_\text{N} < \gamma_{\text{TS}} |& \mathcal{H}_1, k_{\text{L}},k_{\text{M}}) \\ &\leq \exp\left(-k_{\text{L}}D\left( \tilde{\gamma}_{MD}(k_{\text{L}},k_{\text{M}}) || 1 - P_{\text{MD,L}} \right)\right),
        \label{eq:loss_MD_bounds}
\end{aligned}
\end{equation}
where 
\begin{equation}
    \tilde{\gamma}_{\text{MD}}(k_{\text{L}},k_{\text{M}}) := \frac{1}{k_{\text{L}}} \frac{\gamma_{\text{TS}} + k_{\text{M}}w_{0,\text{L}} + k_{\text{L}}w_{0,\text{L}} }{(w_{0,\text{L}} + w_{1,\text{L}})}.
\label{eq:tilde_MD}
\end{equation}

The Chernoff bounds derived in \eqref{eq:loss_FA_bounds} and \eqref{eq:loss_MD_bounds} are  valid  whenever $\tilde{\gamma}_{\text{FA}}(k_{\text{L}},k_{\text{M}}) \in (P_{\text{FA,L}},1)$ and $\tilde{\gamma}_{\text{MD}}(k_{\text{L}},k_{\text{M}}) \in (0,1-P_{\text{MD,L}})$.

From here we upper bound the probability of error corresponding to Case 4 by noticing that our upper bound on the probability of error is maximized when the least legitimate robots are trusted and the most malicious robots are trusted. Define $\underline{k_{\text{L}}} \triangleq \beta_{\text{L}}|\mathcal{L}|+1$ and $\overline{k_{\text{M}}} \triangleq \beta_{\text{M}}|\mathcal{M}|-1$ to be the minimum number of legitimate robots within the region $k_{\text{L}} > \beta_{\text{L}}|\mathcal{L}|$, and the maximum number of malicious robots within the region $k_{\text{M}} < \beta_{\text{M}}|\mathcal{M}|$ that can be trusted, respectively. We formulate this observation in the following lemma.
\begin{lemma}
Consider Case 4 where many legitimate robots are trusted by the FC and few malicious robots are trusted. Without loss of generality, assume $\Pr(\Xi=1) > \Pr(\Xi=0)$. If $\tilde{\gamma}_{\text{FA}}(k_{\text{L}},k_{\text{M}}) \in (P_{\text{FA,L}},1)$ and $\tilde{\gamma}_{\text{MD}}(k_{\text{L}},k_{\text{M}}) \in (0,1-P_{\text{MD,L}})$, then the probability of error for a given $k_{\text{L}}$ and $k_{\text{M}}$ within Case 4 can be upper bound by
\begin{flalign}
&\max_{\substack{k_{\text{L}}> \beta_{\text{L}}|\mathcal{L}|,\\ k_{\text{M}}< \beta_{\text{M}}|\mathcal{M}|}} \overline{p}_{\text{e}}(k_{\text{L}},k_{\text{M}})\nonumber\\
&\leq \Pr(\Xi=0) \max_{\substack{k_{\text{L}}> \beta_{\text{L}}|\mathcal{L}|,\\ k_{\text{M}}< \beta_{\text{M}}|\mathcal{M}|}}P_{\text{FA}}(S_\text{N} \geq \gamma_{\text{TS}} | \mathcal{H}_0, k_{\text{L}},k_{\text{M}}) \nonumber\\ &+ \Pr(\Xi=1) \max_{\substack{k_{\text{L}}> \beta_{\text{L}}|\mathcal{L}|,\\ k_{\text{M}}< \beta_{\text{M}}|\mathcal{M}|}}P_{\text{MD}}(S_\text{N} < \gamma_{\text{TS}} | \mathcal{H}_1, k_{\text{L}},k_{\text{M}}).
\end{flalign}
Additionally, assume $\beta_{\text{L}}|\mathcal{L}| >> \max\{ \beta_{\text{M}}|\mathcal{M}|,1\}$. Then, there exists values $k_{\text{L}}> \beta_{\text{L}}|\mathcal{L}|$, and $k_{\text{M}}< \beta_{\text{M}}|\mathcal{M}|$ such that $\tilde{\gamma}_{\text{FA}}(k_{\text{L}},k_{\text{M}}) \in (P_{\text{FA,L}},1)$ and $\tilde{\gamma}_{\text{MD}}(k_{\text{L}},k_{\text{M}}) \in (0,1-P_{\text{MD,L}})$.

\label{lem:Pe_klkm}
\end{lemma}
\begin{proof}
We start by proving the first part of the lemma, which upper bounds the error probability for Case 4. 
The probability of error for Case 4 and a given $k_{\text{L}}$ and $k_{\text{M}}$ is
\begin{equation}
\begin{aligned}
    \overline{p}_{\text{e}}(k_{\text{L}},k_{\text{M}}) = &\Pr(\Xi=0) P_{\text{FA}}(S_\text{N} \geq \gamma_{\text{TS}} | \mathcal{H}_0, k_{\text{L}},k_{\text{M}}) \\ &+ \Pr(\Xi=1) P_{\text{MD}}(S_\text{N} < \gamma_{\text{TS}} | \mathcal{H}_1, k_{\text{L}},k_{\text{M}}).
\end{aligned}
\end{equation}
The event probabilities $\Pr(\Xi=0)$ and $\Pr(\Xi=1)$ are constant, so in order to upper bound $\overline{p}_{\text{e}}(k_{\text{L}},k_{\text{M}})$, we look to upper bound $P_{\text{FA}}(S_\text{N} \geq \gamma_{\text{TS}} | \mathcal{H}_0, k_{\text{L}},k_{\text{M}})$ and \mbox{$P_{\text{MD}}(S_\text{N} < \gamma_{\text{TS}} | \mathcal{H}_1, k_{\text{L}},k_{\text{M}})$}. We will only derive the result for $P_{\text{FA}}(S_\text{N} \geq \gamma_{\text{TS}} | \mathcal{H}_0, k_{\text{L}},k_{\text{M}})$ since the proof is analogous for $P_{\text{MD}}(S_\text{N} < \gamma_{\text{TS}} | \mathcal{H}_1, k_{\text{L}},k_{\text{M}})$.
\newline \indent
From \eqref{eq:loss_FA_bounds} we see that for every $k_{\text{L}}$ and $k_{\text{M}}$ such that $k_{\text{L}}> \beta_{\text{L}}|\mathcal{L}|$ and $k_{\text{M}}< \beta_{\text{M}}|\mathcal{M}|$ the following holds:
\begin{equation}
\begin{aligned}
    &P_{\text{FA}}(S_\text{N} \geq \gamma_{\text{TS}} | \mathcal{H}_0, k_{\text{L}},k_{\text{M}}) \\
    &\leq \max_{\substack{k_{\text{L}}> \beta_{\text{L}}|\mathcal{L}|,\\ k_{\text{M}}< \beta_{\text{M}}|\mathcal{M}|}}P_{\text{FA}}(S_\text{N} \geq \gamma_{\text{TS}} | \mathcal{H}_0, k_{\text{L}},k_{\text{M}}) \\ 
    &\leq \max_{\substack{k_{\text{L}}> \beta_{\text{L}}|\mathcal{L}|,\\ k_{\text{M}}< \beta_{\text{M}}|\mathcal{M}|}} \exp\left( -k_\text{L} D\left( \tilde{\gamma}_{\text{FA}}(k_{\text{L}},k_{\text{M}}) || P_{\text{FA,L}} \right) \right) \\
    &\stackrel{(a)}{\leq} \exp\left( -\min_{\substack{k_{\text{L}}> \beta_{\text{L}}|\mathcal{L}|,\\ k_{\text{M}}< \beta_{\text{M}}|\mathcal{M}|}}k_\text{L} \cdot\min_{\substack{k_{\text{L}}> \beta_{\text{L}}|\mathcal{L}|,\\ k_{\text{M}}< \beta_{\text{M}}|\mathcal{M}|}}D\left( \tilde{\gamma}_{\text{FA}}(k_{\text{L}},k_{\text{M}}) || P_{\text{FA,L}} \right) \right) \\
    &\stackrel{(b)}{\leq} \exp\left( -\underline{k_\text{L}} \cdot D\left( \tilde{\gamma}_{\text{FA}}(\underline{k_{\text{L}}},\overline{k_{\text{M}}}) || P_{\text{FA,L}} \right) \right),
\end{aligned}
\label{eq:loss_FA_bounds_proof}
\end{equation}
where $(a)$ follows from the nonnegativity of $k_{\text{L}}$ and the KL divergence. The inequality $(b)$ follows by minimizing both terms in the product in $(a)$. The first term is trivially minimized when $k_{\text{L}} = \underline{k_{\text{L}}}$. The KL divergence term attains its minimum at $0$ when $\tilde{\gamma}_{\text{FA}}(k_{\text{L}},k_{\text{M}}) = P_{\text{FA,L}}$. Since $\tilde{\gamma}_{\text{FA}}(k_{\text{L}},k_{\text{M}}) > P_{\text{FA,L}}$, the KL divergence is minimized when $\tilde{\gamma}_{\text{FA}}(k_{\text{L}},k_{\text{M}})$ is minimized. 

From \eqref{eq:tilde_FA} we see that $\tilde{\gamma}_{\text{FA}}(k_{\text{L}},k_{\text{M}})$ is minimized when $k_{\text{M}}$ is maximized, i.e., $k_{\text{M}} = \overline{k_{\text{M}}}$. Now fix $k_{\text{M}} = \overline{k_{\text{M}}}$. Recall the assumption that $\beta_{\text{L}}|\mathcal{L}| >> \max\{ \beta_{\text{M}}|\mathcal{M}|,1\}$, thus $k_{\text{L}} >> k_{\text{M}}$. Therefore, we can rewrite \eqref{eq:tilde_FA} as
\begin{equation}
\begin{aligned}
    \tilde{\gamma}_{\text{FA}}(k_{\text{L}},k_{\text{M}}) &\approx \frac{\gamma_{\text{TS}} + k_{\text{L}}w_{0,\text{L}} }{k_{\text{L}}(w_{0,\text{L}} + w_{1,\text{L}})} \\ &= \frac{\gamma_{\text{TS}}}{k_{\text{L}}(w_{0,\text{L}} + w_{1,\text{L}})} + \frac{w_{0,\text{L}}}{(w_{0,\text{L}} + w_{1,\text{L}})}.
    \end{aligned}
    \label{eq:tilde_gamma_FA_2}
\end{equation}
Since $\Pr(\Xi=1) > \Pr(\Xi=0)$ we have that $\gamma_{\text{TS}} < 0$. Therefore, the expression in \eqref{eq:tilde_gamma_FA_2} is minimized when $k_{\text{L}}$ is minimized, i.e., $k_{\text{L}} = \underline{k_{\text{L}}}$, as long as $\tilde{\gamma}_{\text{FA}}(\underline{k_{\text{L}}},\overline{k_{\text{M}}}) \in (P_{\text{FA,L}},1)$. 

We now proceed to prove the second part of the lemma by showing that the set of values for which $\tilde{\gamma}_{\text{FA}}(\underline{k_{\text{L}}},\overline{k_{\text{M}}}) \in (P_{\text{FA,L}},1)$ is nonempty.

Consider $\underline{k_{\text{L}}} \rightarrow \infty$. Notice that $\tilde{\gamma}_{\text{FA}}(\underline{k_{\text{L}}},\overline{k_{\text{M}}}) \rightarrow \frac{w_{0,\text{L}}}{w_{0,\text{L}} + w_{1,\text{L}}}$. In this case, $\tilde{\gamma}_{\text{FA}}(\underline{k_{\text{L}}},\overline{k_{\text{M}}}) \in (P_{\text{FA,L}},1)$ if $ \frac{w_{0,\text{L}}}{w_{0,\text{L}} + w_{1,\text{L}}} \in (P_{\text{FA,L}},1)$. Since $P_{\text{FA,L}} \in (0,0.5)$ and $P_{\text{MD,L}} \in (0,0.5)$, and $w_{0,\text{L}}, w_{1,\text{L}} > 0$ we have that $\frac{w_{0,\text{L}}}{w_{0,\text{L}} + w_{1,\text{L}}} \in (0,1)$, thus it remains to show that
\begin{equation}
    \frac{w_{0,\text{L}}}{w_{0,\text{L}} + w_{1,\text{L}}} > P_{\text{FA,L}}.
    \label{eq:w0L_w1L_condition}
\end{equation}

We can manipulate \eqref{eq:w0L_w1L_condition} by multiplying both sides by \mbox{$(w_{0,\text{L}} + w_{1,\text{L}})$}, plugging in the expressions for $w_{0,\text{L}}$ and $w_{1,\text{L}}$, and using some algebra to yield
\begin{equation}
    (1-P_{\text{FA,L}}) \log\left( \frac{1-P_{\text{FA,L}}}{P_{\text{MD,L}}} \right) > P_{\text{FA,L}} \log\left( \frac{1-P_{\text{MD,L}}}{P_{\text{FA,L}}} \right).
    \label{eq:w0L_w1L_condition_v3}
\end{equation}
Next, note that
\begin{equation*}
    1 - \frac{1}{x} \leq \log(x) \leq x - 1.
\end{equation*}
Then, we can lower bound the left-hand side of the expression in \eqref{eq:w0L_w1L_condition_v3} and upper bound the right-hand side to give us
\begin{equation*}
    (1-P_{\text{FA,L}})\left(1-\frac{P_{\text{MD,L}}}{1-P_{\text{FA,L}}} \right) > P_{\text{FA,L}} \left( \frac{1-P_{\text{MD,L}}}{P_{\text{FA,L}}} - 1 \right).
\end{equation*}
This reduces to $1 > 1$. Therefore, the condition in \eqref{eq:w0L_w1L_condition} holds for all cases besides when
\begin{equation*}
    1 - \frac{1}{x} = \log(x) = x - 1.
\end{equation*}
This occurs at $x = 1$, which corresponds to
\begin{equation}
    \frac{1-P_{\text{FA,L}}}{P_{\text{MD,L}}} = 1, \quad \frac{1-P_{\text{MD,L}}}{P_{\text{FA,L}}} = 1.
\end{equation}
If we restrict the values of $P_{\text{FA,L}}$ and $P_{\text{MD,L}}$ to $P_{\text{FA,L}} \in (0,0.5]$ and $P_{\text{MD,L}} \in (0,0.5]$ then this corresponds to $P_{\text{FA,L}} = P_{\text{MD,L}} = 0.5$. Since $P_{\text{FA,L}}$ and $P_{\text{MD,L}}$ are bounded away from $0.5$, the condition in \eqref{eq:w0L_w1L_condition} holds for all $P_{\text{FA,L}} \in (0,0.5)$ and $P_{\text{MD,L}} \in (0,0.5)$.
\end{proof}

From \cref{lem:Pe_klkm} we have that
\begin{equation}
    \begin{aligned}
    &\max_{k_{\text{L}}> \beta_{\text{L}}|\mathcal{L}|, k_{\text{M}}< \beta_{\text{M}}|\mathcal{M}|}\overline{p}_{\text{e}}(k_{\text{L}}, k_{\text{M}}) \leq \Pr(\Xi=0) \cdot \\ &e^{\left(-(\beta_{\text{L}}(1-\overline{m})N+1)D\left( \tilde{\gamma}_{FA}(\beta_{\text{L}}(1-\overline{m})N+1,\beta_{\text{M}}\overline{m}N-1) || P_{\text{FA,L}} \right)\right)} \\ &+ \Pr(\Xi=1) \cdot \\ &e^{\left(-(\beta_{\text{L}}(1-\overline{m})N+1)D\left( \tilde{\gamma}_{MD}(\beta_{\text{L}}(1-\overline{m})N+1,\beta_{\text{M}}\overline{m}N-1) || 1 - P_{\text{MD,L}} \right)\right)}.
    \end{aligned}
\end{equation}
Indeed, we see that the upper bound for the probability of error when Case 4 occurs decays exponentially with a rate of
\begin{align*}
(\beta_\text{L}&(1-\overline{m}) N + 1) \cdot \\ \min \{ & D\left( \tilde{\gamma}_{\text{FA}}(\beta_{\text{L}}(1-\overline{m})N+1,\beta_{\text{M}}\overline{m}N-1) || P_{\text{FA,L}} \right), \\ & D\left( \tilde{\gamma}_{\text{MD}}(\beta_{\text{L}}(1-\overline{m})N+1,\beta_{\text{M}}\overline{m}N-1) || 1-P_{\text{MD,L}} \right) \}.
\end{align*}

Since $[\Pr(\text{Case 1}) + \Pr(\text{Case 2})] \rightarrow 0$, $\Pr(\text{Case 3}) \rightarrow 0$, $\Pr(\text{Case 4}) \rightarrow 1$ as $N \rightarrow \infty$, and the upper bound on the probability of error corresponding to Case 4, $\overline{p}_{\text{e}}(\underline{k_{\text{L}}}, \overline{k_{\text{M}}}) \rightarrow 0$, as $N \rightarrow \infty$, and since all upper bounds exhibit exponential decay rates, we conclude that the probability of error decays towards $0$ at least exponentially as the number of robots in the network increases.

\subsection{Analyzing the Limits of the Two Stage Approach}

If the proportion of malicious robots in the network, i.e., $\overline{m}$, is high enough, the probability of error for the Two Stage Approach will plateau. Intuitively, this is due to the fact that if there are too many malicious robots it becomes more beneficial for the FC to guess between $\mathcal{H}_0$ or $\mathcal{H}_1$ using the prior probabilities $\Pr(\Xi = 0)$ and $\Pr(\Xi = 1)$ rather than utilize any measurements from robots. We formally state and prove this observation with the following lemma. Recall that \cref{alg:2stage_discrete} chooses the classification threshold $\gamma_t$ and tie-break probability $p_t$ by computing the probability of error in the presence of a worst-case attack over all values $\hat{\gamma}_t \in \Gamma_t$ and $\hat{p}_t \in \Gamma_p$, where $\Gamma_t = \left\{ \frac{p_{\alpha}(a|t_i=1)}{p_{\alpha}(a|t_i=0)} \right\}_{a\in\mathcal{A}}$, $\Gamma_p = \{0,\delta_p,2\delta_p,\dots,1\}$, and $\delta_p$ is a given discretization constant.

\begin{lemma}
If the worst-case probability of error for every choice of $\hat{\gamma}_t$ and $\hat{p}_t$ is no better than performing event detection with no information, i.e.,
\begin{equation*}
    \overline{P}_{\text{e}}(\hat{\gamma}_t,\hat{p}_t,\overline{m},1,1) \geq \min\{ \Pr(\Xi=0),\Pr(\Xi=1) \},
\end{equation*}
for all $\hat{\gamma}_t \in \Gamma_t$ and all $\hat{p}_t \in \Gamma_p$, then the optimal worst-case probability of error becomes the probability of the less likely event between $\Xi = 0$ and $\Xi = 1$ occurring, i.e.,
\begin{equation*}
    \overline{P}_{\text{e}}(\gamma_t^*,p_t^*,\overline{m},1,1) = \min\{ \Pr(\Xi=0),\Pr(\Xi=1) \}.
\end{equation*}
Furthermore, the Two Stage Approach chooses thresholds $\gamma_t$ and $p_t$ that lead to not trusting any robots, i.e., \mbox{$\gamma_t^* = \max_{a_i \in \alpha} \left\{ \frac{p_{\alpha}(a_i|t_i=1)}{p_{\alpha}(a_i|t_i=0)} \right\}$} and $p_t^* = 0$.
\label{lem:2SA_plateau}
\end{lemma}

\begin{proof}
Let $\hat{\gamma}_t = \max_{a_i \in \alpha} \left\{ \frac{p_{\alpha}(a_i|t_i=1)}{p_{\alpha}(a_i|t_i=0)} \right\}$ and $\hat{p}_t = 0$. This corresponds to the scenario where the measurements from all robots will be discarded by the FC in the first stage. Discarding all measurements simplifies the decision rule in the second stage \eqref{eq:detection_p_e_FC_legitimate_assumption} to
\begin{flalign}
1\underset{\mathcal{H}_{0}}{\overset{\mathcal{H}_{1}}{\begin{smallmatrix}\geqslant\\<\end{smallmatrix}}}\frac{\Pr(\Xi=0)}{\Pr(\Xi=1)}.
\end{flalign}
If $\Pr(\Xi=0) \geq \Pr(\Xi=1)$ then the FC chooses $\mathcal{H}_0$ which leads to an error probability of $\Pr(\Xi=1)$. If $\Pr(\Xi=0) < \Pr(\Xi=1)$ then the FC chooses $\mathcal{H}_1$ which leads to an error probability of $\Pr(\Xi=0)$. Therefore, the probability of error
\begin{equation*}
    \overline{P}_{\text{e}}(\hat{\gamma}_t,\hat{p}_t,\overline{m},1,1) = \min\{ \Pr(\Xi=0),\Pr(\Xi=1) \}.
\end{equation*}
By \cref{alg:2stage_discrete} if the probability of error is greater for all other $\hat{\gamma}_t \in \Gamma_t$ and $\hat{p}_t \in \Gamma_p$, then \mbox{$\gamma_t^* = \max_{a_i \in \alpha} \left\{ \frac{p_{\alpha}(a_i|t_i=1)}{p_{\alpha}(a_i|t_i=0)} \right\}$} and $p_t^* = 0$.
\end{proof}

This lemma formally shows that if at some point the probability of error when trusting any robots is always greater than the probability of error from using the prior probabilities $\Pr(\Xi=0)$ and $\Pr(\Xi=1)$ then \cref{alg:2stage_discrete} chooses $\gamma_t^*$ and $p_t^*$ such that no robots are ever trusted, reducing to the case where the hypothesis prediction is done using the known event probabilities. Let $m^*$ denote the critical proportion of malicious robots that causes the Two Stage Approach to reject all information in the first stage, i.e., for all $\overline{m} \geq m^*$ we have $\overline{P}_{\text{e}}(\gamma_t^*,p_t^*,\overline{m},1,1) = \min\{ \Pr(\Xi=0),\Pr(\Xi=1) \}$.

Next we develop an understanding of how $m^*$ is affected by the quality of the trust values, i.e., as a function of the probability of trusting legitimate and malicious robots, $P_{\text{trust,L}}(\gamma_t,p_t)$ and $P_{\text{trust,M}}(\gamma_t,p_t)$. In order to do so, we assume there is no noise in the sensor measurements of legitimate robots, i.e., $P_{\text{FA,L}} = P_{\text{MD,L}} = 0$. This allows us to simplify the probability of false alarm in \eqref{eq:P_FA_kl_km} by considering $y_i$ to be a deterministic variable with respect to the true legitimacy of robot $i$:
\begin{equation}
    P_{\text{FA}}(S_{\text{N}} \geq \gamma_{\text{TS}} | \mathcal{H}_0) = \Pr(-K_{\text{L}} w_{0,\text{L}} + K_{\text{M}} w_{1,\text{L}} \geq \gamma_{\text{TS}} | \mathcal{H}_0),
    \label{eq:P_FA_mstar}
\end{equation}
where $K_{\text{L}} \in \{ 0,1,\dots,(1-\overline{m})N \}$ and \mbox{$K_{\text{M}} \in \{ 0,1,\dots,\overline{m}N \}$} are random variables that represent the possible number of trusted legitimate and malicious robots, respectively. When $P_{\text{FA,L}} = P_{\text{MD,L}}$ we have that $w_{0,\text{L}} = w_{1,\text{L}}$. Then \eqref{eq:P_FA_mstar} becomes
\begin{equation}
    P_{\text{FA}}(S_{\text{N}} \geq \gamma_{\text{TS}} | \mathcal{H}_0) = \Pr(K_{\text{M}} - K_{\text{L}} \geq 0 | \mathcal{H}_0),
    \label{eq:P_FA_mstar2}
\end{equation}
where we use the fact that $w_{0,\text{L}} \rightarrow \infty$ and $w_{1,\text{L}} \rightarrow \infty$ as $P_{\text{FA,L}} \rightarrow 0$ and $P_{\text{MD,L}} \rightarrow 0$. Similarly, the probability of missed detection becomes
\begin{equation}
    P_{\text{MD}}(S_{\text{N}} < \gamma_{\text{TS}} | \mathcal{H}_1) = \Pr(K_{\text{M}} - K_{\text{L}} > 0 | \mathcal{H}_1).
    \label{eq:P_MD_mstar2}
\end{equation}
The variables $K_{\text{L}}$ and $K_{\text{M}}$ are distributed according to binomial distributions:
\begin{equation}
    K_{\text{L}} \sim \operatorname{BIN}((1-\overline{m})N,P_{\text{trust,L}}), K_{\text{M}} \sim \operatorname{BIN}(\overline{m}N,P_{\text{trust,M}}),
\end{equation}
where $\operatorname{BIN}(n,p)$ corresponds to a binomial distribution with $n$ trials and success probability $p$.
\newline \indent
Define $Z \triangleq K_{\text{M}} - K_{\text{L}}$ to be a discrete random variable corresponding to the difference of the two binomial random variables $K_{\text{M}}$ and $K_{\text{L}}$. We are interested in $\Pr(Z \geq 0)$ for the probability of false alarm, and $\Pr(Z > 0)$ for the probability of missed detection. Then, $m^*$ could be found by finding the minimum $\overline{m}$ such that
\begin{equation}
\begin{aligned}
    \Pr(\Xi=0) \Pr(Z \geq 0) + \Pr(\Xi=1) \Pr(Z > 0) \\ \geq \min\{\Pr(\Xi=0),\Pr(\Xi=1)\},
\end{aligned}
\label{eq:mstar_cond1}
\end{equation}
or equivalently,
\begin{equation}
\begin{aligned}
    \Pr(\Xi=0) \Pr(Z = 0) + \Pr(Z > 0) \\ \geq \min\{\Pr(\Xi=0),\Pr(\Xi=1)\},
\end{aligned}
\end{equation}
where $\Pr(Z \geq 0)$ and $\Pr(Z > 0)$ are a function of $P_{\text{trust,L}}(\gamma_t,p_t)$, $P_{\text{trust,M}}(\gamma_t,p_t)$, $\overline{m}$, and $N$.

When $N$ is large the distribution of $Z$ is approximately normal with mean \[\mu = \overline{m}N(P_{\text{trust,L}}+P_{\text{trust,M}}) - NP_{\text{trust,L}},\] and variance \[\sigma^2 = \overline{m}NP_{\text{trust,M}}(1-P_{\text{trust,M}}) + (1-\overline{m})NP_{\text{trust,L}}(1-P_{\text{trust,L}}).\] The mean is found using the linearity of expectation, and the variance is found by utilizing the fact that the binomial random variables $K_{\text{M}}$ and $K_{\text{L}}$ are conditionally independent given $P_{\text{trust,L}}$, $P_{\text{trust,M}}$, $\overline{m}$, and $N$. Then, we can approximate the probability $\Pr(Z~>~z)$ using the complement distribution function $Q(g) = \frac{1}{\sqrt{2\pi}} \int_g^\infty \exp{-(u^2/2)} du$ where $g = \frac{z-\mu}{\sigma}$. Utilizing this, we have
\begin{equation}
\begin{aligned}
    &\Pr(Z > 0) \approx Q\left( \frac{-\mu}{\sigma} \right) \\ &= \resizebox{0.93\hsize}{!}{$Q\left( \frac{NP_{\text{trust,L}} -\overline{m}N(P_{\text{trust,L}}+P_{\text{trust,M}})}{\sqrt{\overline{m}NP_{\text{trust,M}}(1-P_{\text{trust,M}}) + (1-\overline{m})NP_{\text{trust,L}}(1-P_{\text{trust,L}})}} \right)$},
    \label{eq:MD_approx}
\end{aligned}
\end{equation}
for the probability of missed detection. Similarly, the probability of false alarm $\Pr(Z \geq 0)$ can be upper bound by $\Pr(Z > 0)$. Since $Z$ is the difference of two binomial random variables, the random variable $Z$ is discrete and takes only integer values. However, the probability of false alarm can be lower bound by $\Pr(Z > -1/2)$ using the continuity correction \cite[Ch 4]{devore2015probability}. This lower bound can be approximated as
\begin{equation}
\begin{aligned}
    &\Pr(Z > -1/2) \approx Q\left( \frac{-1/2-\mu}{\sigma} \right) \\ &= \resizebox{0.93\hsize}{!}{$Q\left( \frac{NP_{\text{trust,L}} -\overline{m}N(P_{\text{trust,L}}+P_{\text{trust,M}})-1/2}{\sqrt{\overline{m}NP_{\text{trust,M}}(1-P_{\text{trust,M}}) + (1-\overline{m})NP_{\text{trust,L}}(1-P_{\text{trust,L}})}} \right)$}.
    \label{eq:FA_approx}
\end{aligned}
\end{equation}

\subsubsection{Simulation study for $m^*$}

\begin{figure}[h!]
    \centering
    \includegraphics[scale=0.56]{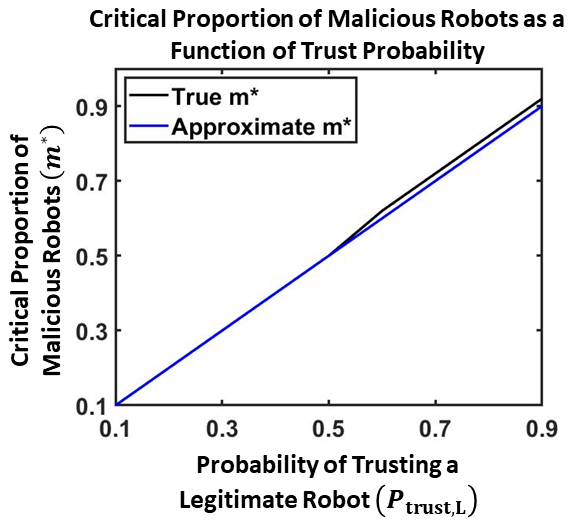}
    \caption[Simulation study to compare the true value for the critical proportion of malicious robots that the Two Stage Approach can handle with the approximated value. It can be seen that the approximate value closely matches the true value, and a fairly linear relationship exists between the probability of trusting legitimate and malicious robots and the critical proportion of malicious robots.]{Simulation study to compare the true $m^*$ value with the approximated value using \eqref{eq:MD_approx} and \eqref{eq:FA_approx} when $N = 50$ and \mbox{$\Pr(\Xi=0)=\Pr(\Xi=1)=0.5$}, for $P_{\text{trust,L}}(\gamma_t,p_t) \in [0.1,0.9]$ with $P_{\text{trust,M}}(\gamma_t,p_t) = 1 - P_{\text{trust,L}}(\gamma_t,p_t)$. It can be seen that the approximate value closely matches the true value, and a fairly linear relationship exists between the probability of trusting legitimate and malicious robots and $m^*$.}
    \label{fig:m_star_sim_study}
\end{figure}

We conclude this section by running a simple simulation study where we compute the true $m^*$ value (referred to as \emph{true $m^*$} in \cref{fig:m_star_sim_study}), found by varying $\overline{m}$ from $0$ to $1$ and choosing the first value such that $\overline{P}_{\text{e}}(\hat{\gamma}_t,\hat{p}_t,\overline{m},1,1) \geq \min\{ \Pr(\Xi=0),\Pr(\Xi=1) \}$. We also compute $m^*$ in the same way while approximating $P_{\text{FA}}(S_{\text{N}} \geq \gamma_{\text{TS}} | \mathcal{H}_0)$ by $\Pr(Z > -1/2)$ in \eqref{eq:FA_approx} and $P_{\text{MD}}(S_{\text{N}} < \gamma_{\text{TS}} | \mathcal{H}_1)$ by $\Pr(Z > 0)$ in \eqref{eq:MD_approx} (referred to as \emph{approximate $m^*$} in \cref{fig:m_star_sim_study}). We compare the results for a case where we set $N = 50$ and $\Pr(\Xi=0)=\Pr(\Xi=1)=0.5$, and vary $P_{\text{trust,L}}(\gamma_t,p_t) \in [0.1,0.9]$ with $P_{\text{trust,M}}(\gamma_t,p_t) = 1 - P_{\text{trust,L}}(\gamma_t,p_t)$. From \cref{fig:m_star_sim_study} it can be seen that our method of approximating the true $m^*$ in \eqref{eq:MD_approx} and \eqref{eq:FA_approx} closely matches the true value. It can also be seen that a fairly linear relationship exists between the probability of trusting legitimate and malicious robots and the critical proportion of malicious robots $m^*$. Moreover, for the case where $\Pr(\Xi=0)=\Pr(\Xi=1)$ this relationship can be estimated by
\begin{equation}
    m^* \approx \frac{P_{\text{trust,L}}(\gamma_t,p_t)}{P_{\text{trust,L}}(\gamma_t,p_t) + P_{\text{trust,L}}(\gamma_t,p_t)}.
\end{equation}

\section{Adversarial Generalized Likelihood Ratio Test}

In this section, we introduce our second approach, called the Adversarial Generalized Likelihood Ratio Test (A-GLRT).The A-GLRT uses both the trust values and measurements simultaneously to arrive at a final decision while estimating the unknown parameters using the maximum likelihood estimation rule. The A-GLRT approach addresses \cref{prob:prob2}.

\subsection{A-GLRT Algorithm} \label{sec:GLRT}
The main purpose of this section is to construct an efficient algorithm that implements the GLRT in \eqref{eq:GLRT_ML_decision_joint}. We can simplify \eqref{eq:GLRT_ML_decision_joint} by recalling that given the true trustworthiness of a robot $t_i$ and the true hypothesis $\mathcal{H}$, the trust value $\alpha_i$ and the measurement $Y_i$ are statistically independent. 
Thus,
\begin{flalign}
&\Pr(\boldsymbol{a},\boldsymbol{y}|\mathcal{H}_1,\boldsymbol{t},P_{\text{MD,M}})\nonumber\\
&=\Pr(\boldsymbol{a}|\mathcal{H}_1,\boldsymbol{t},P_{\text{MD,M}})\Pr(\boldsymbol{y}|\mathcal{H}_1,\boldsymbol{t},P_{\text{MD,M}}),\\
&\Pr(\boldsymbol{a},\boldsymbol{y}|\mathcal{H}_0,\boldsymbol{t},P_{\text{FA,M}})\nonumber\\
&=\Pr(\boldsymbol{a}|\mathcal{H}_0,\boldsymbol{t},P_{\text{FA,M}})\Pr(\boldsymbol{y}|\mathcal{H}_0,\boldsymbol{t},P_{\text{FA,M}}).
\end{flalign}
Furthermore, the trust value $\alpha_i$ is independent of the  true hypothesis $\mathcal{H}$. Thus,
\begin{flalign}
   \Pr(\boldsymbol{a}|\mathcal{H}_1,\boldsymbol{t},P_{\text{MD,M}})=\Pr(\boldsymbol{a}|\mathcal{H}_0,\boldsymbol{t},P_{\text{FA,M}})=\Pr(\boldsymbol{a}|\boldsymbol{t}). 
\end{flalign}

Hence, we obtain
\begin{flalign}\label{eq:GLRT_ML_decision_independent}
\frac{\max\limits_{\boldsymbol{t}\in\{0,1\}^N,P_{\text{MD,M}}\in[0,1]}\Pr(\boldsymbol{a}|\boldsymbol{t})\Pr(\boldsymbol{y}|\mathcal{H}_1,\boldsymbol{t},P_{\text{MD,M}})}{\max\limits_{\boldsymbol{t}\in\{0,1\}^N,P_{\text{FA,M}}\in[0,1]}\Pr(\boldsymbol{a}|\boldsymbol{t})\Pr(\boldsymbol{y}|\mathcal{H}_0,\boldsymbol{t},P_{\text{FA,M}})}\underset{\mathcal{H}_{0}}{\overset{\mathcal{H}_{1}}{\gtrlesseqslant}} \gamma_{\text{GLRT}}.
\end{flalign}
We choose $\gamma_{\text{GLRT}}=\frac{\Pr(\Xi=0)}{\Pr(\Xi=1)}$ since we do not assume anything about the the prior distribution of $\boldsymbol{t}$. 
The challenging part of using the GLRT in this problem is calculating the maximum likelihood estimations for both the numerator and denominator. The unknown $\boldsymbol{t}$ is a discrete multidimensional variable while $P_{\text{MD,M}}$ and $P_{\text{FA,M}}$ are continuous variables restricted to the domain $[0,1]$. Therefore, calculating the MLE is not trivial. Due to symmetry in the calculation of the numerator and denominator in \eqref{eq:GLRT_ML_decision_independent}, we focus our discussion on the calculation of the numerator. 

Using Assumption~\ref{assumption:trust_val} about the trust values, we obtain the following formulation of $\Pr(\boldsymbol{a}|\boldsymbol{t})$:
$$\Pr(\boldsymbol{a}|\boldsymbol{t}) = \prod_{i=1}^Np_{\alpha}(a_i|t_i).$$ Additionally, we obtain the following equations using the i.i.d. assumption about measurements: 
\begin{flalign}
    & \Pr(\boldsymbol{y}|\mathcal{H}_1,\boldsymbol{t},P_{\text{MD,M}})=\prod_{i:t_i=1} (1-P_{\text{MD,L}})^{y_i}\cdot P_{\text{MD,L}}^{1-y_i}\nonumber\\
   &\hspace{3cm}\cdot \prod_{i:t_i=0} (1-P_{\text{MD,M}})^{y_i}\cdot P_{\text{MD,M}}^{1-y_i}, \label{eq:assumption_y_given_h1_t} \\
   & \Pr(\boldsymbol{y}|\mathcal{H}_0,\boldsymbol{t},P_{\text{FA,M}})=\prod_{i:t_i=1} P_{\text{FA,L}}^{y_i}\cdot(1-P_{\text{FA,L}})^{1-y_i} \nonumber\\
   &\hspace{3cm}\cdot \prod_{i:t_i=0} P_{\text{FA,M}}^{y_i}\cdot(1-P_{\text{FA,M}})^{1-y_i}\label{eq:assumption_y_given_h0_t}.
\end{flalign}
Using these equations, we write the numerator as:
\begin{equation}
    \begin{aligned}
    \label{eq:GLRT_numerator_maximization}
    \max_{\boldsymbol{t}\in\{0,1\}^N,P_{\text{MD,M}}\in[0,1]}\left\{\prod_{i:t_i=1}p_{\alpha}(a_i|t_i)P_{\text{MD,L}}^{1-y_i}(1-P_{\text{MD,L}})^{y_i} \right. \cdot \\ 
    \left. \prod_{i:t_i=0}p_{\alpha}(a_i|t_i)P_{\text{MD,M}}^{1-y_i}(1-P_{\text{MD,M}})^{y_i} \right\}.
    \end{aligned}
\end{equation}

Since there is no clear way to optimize \eqref{eq:GLRT_numerator_maximization} over variables $\boldsymbol{t}$ and $P_{\text{MD,M}}$ at the same time, we reformulate the problem as two nested optimizations using the Principle of Iterated Suprema \cite[p. 515]{olmsted1959real}, that is: 
$$
\begin{aligned}
    \sup\{f(z,w): z \in \mathcal{Z}, w \in \mathcal{W}\} =
    \sup_{z \in \mathcal{Z}}\{\sup_{w \in \mathcal{W}}\{f(z,w)\}\} \\
    = \sup_{w \in \mathcal{W}}\{\sup_{z \in \mathcal{Z}}\{f(z,w)\}\},
    \end{aligned}
$$
where $f \colon \mathcal{Z}\times \mathcal{W} \to \mathbb{R}$, and $\mathcal{Z},\mathcal{W}\subseteq \mathbb{R}^d$. By the Principle of Iterated Suprema we can calculate the maximization in \eqref{eq:GLRT_numerator_maximization} in two different ways. 
We rewrite the maximization problem as:
\begin{equation}
    \begin{aligned}\label{eq:GLRT_numerator_t_first}
    \max_{\boldsymbol{t}\in\{0,1\}^N}\left\{\max_{P_{\text{MD,M}}\in[0,1]}\left\{\prod_{i:t_i=1}p_{\alpha}(a_i|t_i)P_{\text{MD,L}}^{1-y_i}(1-P_{\text{MD,L}})^{y_i} \right. \right. \cdot \\
    \left. \left.\prod_{i:t_i=0}p_{\alpha}(a_i|t_i)P_{\text{MD,M}}^{1-y_i}(1-P_{\text{MD,M}})^{y_i}\right\}\right\}.
    \end{aligned}
\end{equation}

With this formulation, one possible way to calculate the maximization is iterating over all vectors  $\boldsymbol{t}$ in the set $\{0,1\}^N$; then for each $\boldsymbol{t}$, calculating the inner maximization. We show how to calculate the inner maximization in the following lemma.

\begin{lemma}
\label{lem:lemma_t_given}
Let $\boldsymbol{t}$ and $\boldsymbol{y}$ be given vectors in $\{0,1\}^N$. Assume that $p_{\alpha}(a_i|t_i)$ is known both $t_i=0$ and $t_i=1$, and that $\sum_{i:t_i=0}1>0$. Then,
\begin{equation}
    \begin{aligned}
    &\prod_{i:t_i=1}p_{\alpha}(a_i|t_i)P_{\text{\emph{MD,L}}}^{1-y_i}(1-P_{\text{\emph{MD,L}}})^{y_i} \cdot\\
   & \qquad\prod_{i:t_i=0}p_{\alpha}(a_i|t_i)P_{\text{\emph{MD,M}}}^{1-y_i}(1-P_{\text{\emph{MD,M}}})^{y_i}
    \end{aligned}
\label{eq:numerator_inner_max}
\end{equation}

is maximized by $\widehat{P}_{\text{\emph{MD,M}}}=\frac{\sum_{i:t_i=0}(1-y_i)}{\sum_{i:t_i=0}1}$.
Additionally, if $\sum_{i:t_i=0}1=0$, i.e., $|\{i:t_i=0\}|=0$, any choice $\widehat{P}_{\text{\emph{MD,M}}}\in[0,1]$ maximizes \eqref{eq:numerator_inner_max}.
\end{lemma}
\begin{proof}
First, observe that given the vector $\boldsymbol{t}$,  \eqref{eq:numerator_inner_max} is maximized by MLE of $\prod_{i:t_i=0}p_{\alpha}(a_i|t_i)P_{\text{MD,M}}^{1-y_i}(1-P_{\text{MD,M}})^{y_i}$. 
Furthermore, since
\begin{flalign}
&\prod_{i:t_i=0}p_{\alpha}(a_i|t_i)P_{\text{MD,M}}^{1-y_i}(1-P_{\text{MD,M}})^{y_i}\nonumber\\
&=\left(\prod_{i:t_i=0}p_{\alpha}(a_i|t_i)\right)\left(\prod_{i:t_i=0}P_{\text{MD,M}}^{1-y_i}(1-P_{\text{MD,M}})^{y_i}\right),   
\end{flalign}
it follows that  \eqref{eq:numerator_inner_max} is maximized by the MLE of $\prod_{i:t_i=0}P_{\text{MD,M}}^{1-y_i}(1-P_{\text{MD,M}})^{y_i}$.
\\
This is a well-known estimation problem \cite[Problem 7.8]{kay1993estimation}, that together with the invariance property of the MLE \cite[Theorem 7.2]{kay1993estimation} leads to the optimal estimator 
\[\widehat{P}_{\text{MD,M}}=\frac{\sum_{i:t_i=0}(1-y_i)}{\sum_{i:t_i=0}1}.\]
Note, that this estimator is equal to the empirical missed detection probability of the measurements sent by the malicious robots.
Finally, it is easy to validate that if  $|\{i:t_i=0\}|=0$, any choice of $\widehat{P}_{\text{MD,M}}\in[0,1]$ maximizes \eqref{eq:numerator_inner_max}.
\end{proof}

Unfortunately, since the set $\{0,1\}^N$ grows exponentially with the number of robots in the network, this approach is computationally intractable for large robot networks. Therefore, we look for an alternative solution. Another equivalent formulation of the maximization problem in \eqref{eq:GLRT_numerator_maximization} that is obtained by the Principle of Iterated Supremum is
\begin{equation}
    \begin{aligned}\label{eq:GLRT_numerator_pma_first}
    \max_{P_{\text{MD,M}}\in[0,1]}\left\{\max_{\boldsymbol{t}\in\{0,1\}^N}\left\{\prod_{i:t_i=1}p_{\alpha}(a_i|t_i)P_{\text{MD,L}}^{1-y_i}(1-P_{\text{MD,L}})^{y_i} \right. \right. \cdot \\
    \left. \left.\prod_{i:t_i=0}p_{\alpha}(a_i|t_i)P_{\text{MD,M}}^{1-y_i}(1-P_{\text{MD,M}})^{y_i}\right\}\right\},
    \end{aligned}
\end{equation}
where the order of variables that the maximization is taken over is flipped. Since the variable $P_{\text{MD,M}}$ belongs to an uncountably infinite set, it is impossible to perform the maximization with this formulation. However, assuming that we have a given $P_{\text{MD,M}}$, the inner maximization can still be calculated. The following lemma shows how to calculate the inner maximization.

\begin{lemma}
\label{lem:lemma_P_given}
Let $P_{\text{\emph{MD,M}}}$, $\boldsymbol{a}$, and $\mathbf{y}$ be given. Additionally, assume that $p_{\alpha}(a_i|t_i)$ is known for both $t_i=0$ and $t_i=1$.
Let 
$$
\begin{aligned}
    c_{\text{\emph{L}},i}=p_{\alpha}(a_i|t_i)P_{\text{\emph{MD,L}}}^{1-y_i}(1-P_{\text{\emph{MD,L}}})^{y_i}
\end{aligned}
$$ and 
$$
\begin{aligned}
    c_{\text{\emph{M}},i}= p_{\alpha}(a_i|t_i)P_{\text{\emph{MD,M}}}^{1-y_i}(1-P_{\text{\emph{MD,M}}})^{y_i}.
\end{aligned}
$$ 
If the estimated robot identity vector $\hat{\boldsymbol{t}}$ is constructed by choosing $\hat{t_i}=1$ if $c_{\text{\emph{L}},i}\geq c_{\text{\emph{M}},i}$ and $\hat{t_i}=0$ otherwise, where $\hat{t_i}$ is the $i^{th}$ component of $\hat{\boldsymbol{t}}$, then, $\hat{\boldsymbol{t}}$ is a vector that maximizes the expression \eqref{eq:numerator_inner_max}. Moreover, maximization with this approach requires $\mathcal{O}(N)$ comparisons.
\end{lemma}
\begin{proof}
First, we reformulate \eqref{eq:numerator_inner_max} as:
    \begin{flalign}\label{eq:numerator_inner_max_over_all_robots}
    &\prod_{i=1}^{N}(p_{\alpha}(a_i|t_i)P_{\text{MD,L}}^{1-y_i}(1-P_{\text{MD,L}})^{y_i})^{t_i} \cdot\nonumber\\ 
    &\qquad(p_{\alpha}(a_i|t_i)P_{\text{MD,M}}^{1-y_i}(1-P_{\text{MD,M}})^{y_i})^{1-t_i},
    \end{flalign}
where the product is calculated by going through all robots rather than going through legitimate and malicious robots separately. 
We define $$c_{\text{L},i}=p_{\alpha}(a_i|t_i)P_{\text{MD,L}}^{1-y_i}(1-P_{\text{MD,L}})^{y_i},$$ and $$c_{M,i}= p_{\alpha}(a_i|t_i)P_{\text{MD,M}}^{1-y_i}(1-P_{\text{MD,M}})^{y_i}.$$ Then, the expression in \eqref{eq:numerator_inner_max_over_all_robots} becomes:
    \begin{flalign}
    \label{eq:numerator_inner_max_over_all_robots_simplified}
    &\prod_{i=1}^{N}c_{\text{L},i}^{t_i} \cdot c_{\text{M},i}^{1-t_i}.
    \end{flalign}
Let $0\log{0}=1$, thus $0^0=1$. Then, the expression \eqref{eq:numerator_inner_max_over_all_robots_simplified} is maximized when choosing $t_i=1$ if $c_{L,i}\geq c_{M,i}$ and $t_i=0$ otherwise. Since this comparison needs to be performed for every $i \in \mathcal{N}$, $\mathcal{O}(N)$ comparisons need to be performed.
\end{proof}
Now, we consider these two perspectives together to introduce an efficient calculation of the numerator of the GLRT given in \eqref{eq:GLRT_numerator_maximization}. By Lemma~\ref{lem:lemma_t_given}, we can see that the optimum value of $P_{\text{MD,M}}$ has a special structure. Exploiting this knowledge, we can restrict the set that $P_{\text{MD,M}}$ belongs to in \eqref{eq:GLRT_numerator_pma_first}. Then, the inner maximization can be calculated using Lemma~\ref{lem:lemma_P_given}. The following theorem builds on this intuition to provide an efficient calculation of \eqref{eq:GLRT_numerator_maximization}.

\begin{theorem}
Assume that $(\boldsymbol{t}^*,P_{\text{MD,M}}^*)$ attains the maximization in \eqref{eq:GLRT_numerator_maximization}. Then, for each vector of measurements $\mathbf{y}$ and trust values $\mathbf{a}$, $P_{\text{MD,M}}^*$ belongs to the set $\mathcal{P}$ where
$$\mathcal{P}\triangleq\left\{\frac{T_n}{T_d}\right\}_{T_n\in\{0,\ldots,T_d\},T_d\in\{1,\ldots,N\}},$$
and $|\mathcal{P}|\leq N^2+1$. Moreover, the maximization in \eqref{eq:GLRT_numerator_maximization} can be calculated by iterating over $\mathcal{O}(N^2)$ different values in $\mathcal{P}$ and performing $\mathcal{O}(N)$ comparisons.
\label{thm:efficient_maximization}
\end{theorem}

\begin{proof}
First, we will approach the problem by rewriting it as \eqref{eq:GLRT_numerator_pma_first} using the Principle of Iterated Suprema:
$$
\begin{aligned}
\max_{P_{\text{MD,M}}\in[0,1]}\left\{\max_{\boldsymbol{t}\in\{0,1\}^N}\left\{\prod_{i:t_i=1}p_{\alpha}(a_i|t_i)P_{\text{MD,L}}^{1-y_i}(1-P_{\text{MD,L}})^{y_i} \right. \right. \cdot \\
\left. \left.\prod_{i:t_i=0}p_{\alpha}(a_i|t_i)P_{\text{MD,M}}^{1-y_i}(1-P_{\text{MD,M}})^{y_i}\right\}\right\},
\end{aligned}
$$
By Lemma~\ref{lem:lemma_P_given}, we can calculate the inner maximization for a given $P_{\text{MD,M}}$. Notice that, since the calculation requires a comparison for each robot, $\mathcal{O}(N)$ comparisons need to be performed for this maximization. Now, consider the other formulation of the problem given by \eqref{eq:GLRT_numerator_t_first}. From Lemma~\ref{lem:lemma_t_given}, we can see that the optimal $P_{\text{MD,M}}$ only depends on the number of ones and zeros of malicious robots for a given $\boldsymbol{t}$. Moreover, the permutation of ones and zeros of malicious robots for a given $\boldsymbol{t}$ does not change the optimum and only the total number of ones and zeros does. We will restrict the set that the outer maximization process iterates over in \eqref{eq:GLRT_numerator_pma_first} based on this observation.
\\
Denote $$\mathcal{P}\triangleq\left\{\frac{T_n}{T_d}\right\}_{T_n\in\{0,\ldots,T_d\},T_d\in\{1,\ldots,N\}},$$
and observe that $|\mathcal{P}|\leq N^2+1$.
It follows from the Lemma~\ref{lem:lemma_t_given} that for each value $\boldsymbol{t}$ in the outer maximization of \eqref{eq:GLRT_numerator_t_first}, except the case where $\boldsymbol{t}$ consist of all ones, the optimal value of $P_{\text{MD,M}}$ belongs to the set $\mathcal{P}$. Moreover, in the case where $\boldsymbol{t}$ consists of all ones, any choice of $P_{\text{MD,M}}$ maximizes the expression. Hence, without loss of generality, it suffices to look for an optimizer $P_{\text{MD,M}}$ of \eqref{eq:GLRT_numerator_t_first} in the set $\mathcal{P}$.
Therefore, there are only  $\mathcal{O}(N^2)$ possible values that the optimal $P_{\text{MD,M}}$ can take. Thus, we can reformulate \eqref{eq:GLRT_numerator_pma_first} as:
$$
\begin{aligned}
\max_{P_{\text{MD,M}}\in\mathcal{P}}\left\{\max_{\boldsymbol{t}\in\{0,1\}^N}\left\{\prod_{i:t_i=1}p_{\alpha}(a_i|t_i)P_{\text{MD,L}}^{1-{y_i}}(1-P_{\text{MD,L}})^{{y_i}} \right. \right. \cdot \\
\left. \left.\prod_{i:t_i=0}p_{\alpha}(a_i|t_i)P_{\text{MD,M}}^{1-{y_i}}(1-P_{\text{MD,M}})^{{y_i}}\right\}\right\}.
\end{aligned}
$$
Therefore, this maximization can be calculated by iterating over $\mathcal{O}(N^2)$ different values of $P_{\text{MD},\text{M}}$ and for each value, performing $\mathcal{O}(N)$ comparisons. A similar approach can be adapted for calculating the denominator as well.
\end{proof}
Now, using Theorem~\ref{thm:efficient_maximization}, we introduce the algorithm A-GLRT, which makes a decision based on the GLRT given by \eqref{eq:GLRT_ML_decision_independent}.\\
\begin{corollary}
The GLRT given by \eqref{eq:GLRT_ML_decision_independent} can be calculated by Algorithm~\ref{alg:A-GLRT} which is referred to as the A-GLRT algorithm. The A-GLRT algorithm requires $\mathcal{O}(N^3)$ comparisons.
\end{corollary}
\begin{proof}
Calculation of the maximization in the numerator can be calculated in $\mathcal{O}(N^2)$ iterations and performing $\mathcal{O}(N)$ comparisons at each iteration as described by Theorem~\ref{thm:efficient_maximization}. Therefore, it requires $\mathcal{O}(N^3)$ comparisons in total. Similarly, maximization of the denominator requires the same amount of computation and can be calculated in a similar manner using $P_{\text{FA,M}}$ instead of $P_{\text{MD,M}}$. After that, a final comparison is made by comparing the ratio of the numerator and denominator with $\gamma_{\text{GLRT}}=\frac{\Pr(\Xi=0)}{\Pr(\Xi=1)}$. Algorithm~\ref{alg:A-GLRT} follows these steps, therefore, it requires $\mathcal{O}(N^3)$ comparisons in total.
\end{proof}

\begin{algorithm}[h!]
\caption{A-GLRT \\ Input: $\mathbf{y}$, $\mathbf{a}$, $P_{\text{FA,L}}$, $P_{\text{MD,L}}$, $\Pr(\Xi=0)$, $\Pr(\Xi=1)$, $p_{\alpha}(a_i|t=~1)$, $p_{\alpha}(a_i|t=0)$, N \\ Output: Decision $\mathcal{H}_0$ or $\mathcal{H}_1$}
\label{alg:A-GLRT}
\begin{algorithmic}[1]

\State Set $\mathcal{P} = \left\{\frac{T_n}{T_d}\right\}_{T_n\in\{0,\ldots,T_d\},T_d\in\{1,\ldots,N\}}$.

\State Set $\gamma_{\text{GLRT}} = \frac{\Pr(\Xi=0)}{\Pr(\Xi=1)}$.

\State Set $l_\text{num,max} = 0, l_\text{denom,max} = 0$.

\ForAll{$P_{M} \in \mathcal{P}$}

\State Set $P_{\text{MD,M}} = P_{M}, P_{\text{FA,M}} = P_{M}$.

\State Set $l_\text{num} = 1, l_\text{denom} = 1$.

\For{i=1 to N}

\State Set $c_{\text{L},i}=p_{\alpha}(a_i|t_i=1)P_{\text{MD,L}}^{1-y_i}(1-P_{\text{MD,L}})^{y_i}$.

\State Set $c_{\text{M},i}=p_{\alpha}(a_i|t_i=0)P_{\text{MD,M}}^{(1-y_i)}(1-P_{\text{MD},\text{M}})^{y_i}$.

\If {$c_{\text{L},i} \geq c_{\text{M},i}$}

\State Set $l_\text{num} = l_\text{num}\cdot c_{\text{L},i}$.

\Else

\State Set $l_\text{num} = l_\text{num}\cdot c_{\text{M},i}$.

\EndIf

\EndFor

\If {$l_\text{num} > l_\text{num,max}$}

\State Set $l_\text{num,max} = l_\text{num}$.

\EndIf

\State Repeat steps 7-15 for the denominator.

\EndFor

\If {$\frac{l_\text{num,max}}{l_\text{denom,max}} > \gamma_{\text{GLRT}}$}
    \State Return decision $\mathcal{H}_1$ 
\Else
    \State Return decision $\mathcal{H}_0$ 
\EndIf

\end{algorithmic}
\end{algorithm}
Finally, we investigate how the measurements $\boldsymbol{y}$ and stochastic trust values $\boldsymbol{\alpha}$ are being used by the A-GLRT algorithm. Considering \eqref{eq:numerator_inner_max_over_all_robots}, an equivalent decision rule to the one derived in Lemma~\ref{lem:lemma_P_given} is given as:
\begin{flalign}\label{eq:lemma_P_given_alternative_formulation_num}
\frac{p_{\alpha}(a_i|t_i=1)}{p_{\alpha}(a_i|t_i=0)}\underset{\hat{t}_i=0}{\overset{\hat{t}_i=1}{\begin{smallmatrix}\geqslant\\<\end{smallmatrix}}} \frac{P_{\text{MD,M}}^{1-y_i}(1-P_{\text{MD,M}})^{y_i}}{P_{\text{MD,L}}^{1-y_i}(1-P_{\text{MD,L}})^{y_i}}.
\end{flalign}

With this new perspective, we can gain more insights about the A-GLRT. First, we can see that the A-GLRT is essentially performing a likelihood ratio test with trust values for each robot to decide if they are legitimate or not using different threshold values based on the measurement coming from that robot. For now, let's assume that $P_{\text{MD,M}}$ is not 0 or 1. Then, we can see that as trust values become more accurate, meaning that the ratio $\frac{p_{\alpha}(a_i|t_i=1)}{p_{\alpha}(a_i|t_i=0)}$ approaches infinity if $t_i=1$ or approaches zero otherwise, for all values that $\alpha_i$ can take, the finite threshold value becomes insignificant and the decision is made using trust values only. This situation agrees with the intuition as well since trust values would become true indicators of robot identities. In the next section, we formalize this intuition.

\subsection{Behavior of the A-GLRT as the Quality of the Trust Values Increase}
\label{subsec:quality_of_alpha_GLRT}
In this section, we characterize the behavior of the A-GLRT algorithm as the quality of the trust values increase. For the rest of this section only, we focus on the special case where $\alpha_i$ is a discrete random variable drawn from a Bernoulli distribution:
\begin{assumption}
\label{assumption:bernouilli}
Let the vector $\boldsymbol{t}$ denote the true identities of the robots in the network. We assume that the distribution of $\alpha_i$ when robot $i$ is legitimate is the Bernoulli distribution with probability $1-p_{e,1}$,
\begin{equation}
    p_{\alpha}(a_i | t_i = 1) \sim ~ \text{Bernoulli}(1-p_{e,1}).
\end{equation}
Similarly, we assume that the distribution given that robot $i$ is malicious is the Bernoulli distribution with probability $p_{e,0}$,
\begin{equation}
    p_{\alpha}(a_i | t_i = 0) \sim ~ \text{Bernoulli}(p_{e,0}).
\end{equation}
\end{assumption}
Under this assumption, we are interested in the case where the conditional expectation in the case that $i\in~\mathcal{M}$ approaches 0 such that $\mathbb{E}[\alpha_i | t_i = 0] \to 0$, and the conditional in the case that $i\in \mathcal{L}$ approaches 1 such that $\mathbb{E}[\alpha_i | t_i = 1] \to 1$. Since the expected value of $\text{Bernoulli}(p)$ is equal to $p$, a direct implication of this limit behavior and Assumption~\ref{assumption:bernouilli} is that both $p_{e,1}$ and $p_{e,0}$ approach 0. 
Let $(\boldsymbol{t_n}^*,P_{MD,M}^*)$ and $(\boldsymbol{t_d}^*,P_{FA,M}^*)$ be maximizers of the numerator and denominator in the GLRT decision rule \eqref{eq:GLRT_ML_decision_independent}
$$
\frac{\max\limits_{\boldsymbol{t}\in\{0,1\}^N,P_{\text{MD,M}}\in[0,1]}\Pr(\boldsymbol{a}|\boldsymbol{t})\Pr(\boldsymbol{y}|\mathcal{H}_1,\boldsymbol{t},P_{\text{MD,M}})}{\max\limits_{\boldsymbol{t}\in\{0,1\}^N,P_{\text{FA,M}}\in[0,1]}\Pr(\boldsymbol{a}|\boldsymbol{t})\Pr(\boldsymbol{y}|\mathcal{H}_0,\boldsymbol{t},P_{\text{FA,M}})}\underset{\mathcal{H}_{0}}{\overset{\mathcal{H}_{1}}{\gtrlesseqslant}} \gamma_{\text{GLRT}},
$$ respectively: We want to show that both $\boldsymbol{t_n}^*$ and $\boldsymbol{t_d}^*$ are equal to the true trust vector $\boldsymbol{t}$ with high probability. Moreover, when these estimated vectors are equal to each other, i.e.,  $\boldsymbol{t_n}^*=\boldsymbol{t_d}^*$, A-GLRT is equivalent to the likelihood ratio test using the measurements of legitimate robots only.

\begin{lemma}
Assume that \cref{assumption:bernouilli} holds. Let $(\boldsymbol{t_n}^*,P_{\text{MD,M}}^*)$ and $(\boldsymbol{t_d}^*,P_{\text{FA,M}}^*)$ be maximizers of the numerator and denominator in the GLRT decision rule \eqref{eq:GLRT_ML_decision_independent}, respectively. Then, both $\boldsymbol{t_n}^*$ and $\boldsymbol{t_d}^*$ are equal to $\boldsymbol{t}$ with high probability given that $\mathbb{E}[\alpha_i | t_i = 0] \to 0$ for every $i\in\mathcal{M}$ and $\mathbb{E}[\alpha_i | t_i = 1] \to 1$ for every $i\in\mathcal{L}$. 
\label{lem:t_correctly_estimated}
\end{lemma}
\begin{proof}
We show the proof only for the numerator for conciseness. However, a symmetric argument applies to the denominator as well. Moreover, we drop the subscript in $\boldsymbol{t_n}^*$ for readability and instead we denote it with $\boldsymbol{t^*}$. We want to show that the probability $\Pr(\boldsymbol{t^*}\neq\boldsymbol{t}|\mathcal{H}_1,\boldsymbol{t}, P_{\text{MD,M}}^*)$ goes to zero as $\mathbb{E}[\alpha_i | t_i = 0] \to 0$ and $\mathbb{E}[\alpha_i | t_i = 1] \to 1$. Our strategy is to split this probability into two cases using the law of total probability: the first case is the case where the vector of trust values do not match the true trust vector, i.e., $\boldsymbol{a}\neq\boldsymbol{t}$ and the second case where $\boldsymbol{a}=\boldsymbol{t}$. The intuition is that the probability of the first case goes to zero, and $\boldsymbol{t^*}$ will be equal to $\boldsymbol{t}$ with high probability in the second case since $p_{e,1}, p_{e,0} \to 0$. 
Now, we will show this formally.
\begin{equation}
    \begin{aligned}
        &\Pr(\boldsymbol{t^*}\neq\boldsymbol{t}|\mathcal{H}_1,\boldsymbol{t}, P_{\text{MD,M}}^*) = \\
&\Pr(\boldsymbol{t^*}\neq\boldsymbol{t}|\boldsymbol{a}\neq\boldsymbol{t}, \boldsymbol{t},\mathcal{H}_1,P_{\text{MD,M}}^*)\Pr(\boldsymbol{a}\neq\boldsymbol{t}|\boldsymbol{t}) \\ 
&\Pr(\boldsymbol{t^*}\neq\boldsymbol{t}|\boldsymbol{a}=\boldsymbol{t}, \boldsymbol{t},\mathcal{H}_1,P_{\text{MD,M}}^*)\Pr(\boldsymbol{a}=\boldsymbol{t}|\boldsymbol{t}).
    \end{aligned}
\label{eq:A-GLRT_mistake_total_law}
\end{equation}
We can bound the probability $\Pr(\boldsymbol{a}\neq\boldsymbol{t}|\boldsymbol{t})$ as
\begin{equation}
    \begin{aligned}
    \Pr(\boldsymbol{a}\neq\boldsymbol{t}|\boldsymbol{t}) &=
    \Pr\left(\bigcup_{i\in\mathcal{L}\cup\mathcal{M}}\{a_i \neq t_i\}|\boldsymbol{t}\right) \\ &\leq
    \sum_{i\in\mathcal{N}}\Pr(a_i\neq t_i|t_i) \\ &= 
    |\mathcal{M}|p_{e,0} + |\mathcal{L}|p_{e,1} 
    \end{aligned}
\end{equation}
Since $p_{e,1},p_{e,0} \to 0$, $\Pr(\boldsymbol{a}\neq\boldsymbol{t}|\boldsymbol{t})$ goes to $0$ and the first term in \eqref{eq:A-GLRT_mistake_total_law} vanishes. Now let's consider the second term. We want to show that the probability \mbox{$\Pr(\boldsymbol{t^*}\neq\boldsymbol{t}|\boldsymbol{a}=\boldsymbol{t}, \boldsymbol{t},\mathcal{H}_1,P_{\text{MD,M}}^*)$} goes to $0$. For contradiction, assume that $\boldsymbol{t^*}\neq\boldsymbol{t}$. Remember that $(\boldsymbol{t}^*,P_{\text{MD,M}}^*)$ maximize the numerator by definition. The numerator is calculated as:
\begin{equation}
    \begin{aligned}
    \prod_{i:t_i=1}p_{\alpha}(a_i|t_i^*)P_{\text{MD,L}}^{1-y_i}(1-P_{\text{MD,L}})^{y_i} \cdot \\
   \prod_{i:t_i=0}p_{\alpha}(a_i|t_i^*)P_{\text{MD,M}}^{*1-y_i}(1-P_{\text{MD,M}}^*)^{y_i}.
    \end{aligned}
\end{equation}
Since $\boldsymbol{t^*}\neq\boldsymbol{t}$ and $\boldsymbol{a}=\boldsymbol{t}$, we have 
\begin{equation}
    \begin{aligned}
    \prod_{i:t_i^*=1}p_{\alpha}(a_i|t_i^*)P_{\text{MD,L}}^{1-y_i}(1-P_{\text{MD,L}})^{y_i} \cdot
    \\
    \prod_{i:t_i^*=0}p_{\alpha}(a_i|t_i^*)P_{\text{MD,M}}^{*1-y_i}(1-P_{\text{MD,M}}^*)^{y_i} \leq \max{(p_{e,1},p_{e,0})}.
    \end{aligned}
\end{equation}
Now, let $(\boldsymbol{t},\hat{P}_{\text{MD,M}})$ be another pair of estimators for the numerator where $\hat{P}_{\text{MD,M}}=0.5$. Using this pair of estimators, we can calculate the numerator as
\begin{equation}
    \begin{aligned}
    \prod_{i:t_i=1}p_{\alpha}(a_i|t_i)P_{\text{MD,L}}^{1-y_i}(1-P_{\text{MD,L}})^{y_i} \cdot
    \\
    \prod_{i:t_i=0}p_{\alpha}(a_i|t_i)\hat{P}_{\text{MD,M}}^{1-y_i}(1-\hat{P}_{\text{MD,M}})^{y_i} = \\
    (\frac{1-p_{e,0}}{2})^{|\mathcal{M}|} \cdot \prod_{i:t_i=1}(1-p_{e,1})P_{\text{MD,L}}^{1-y_i}(1-P_{\text{MD,L}})^{y_i}.
    \end{aligned}
\end{equation}
Since $p_{e,1}, p_{e,0} \to 0$, we have
\begin{equation}
    \begin{aligned}
    \prod_{i:t_i^*=1}p_{\alpha}(a_i|t_i^*)P_{\text{MD,L}}^{1-y_i}(1-P_{\text{MD,L}})^{y_i} \cdot
    \\
    \prod_{i:t_i^*=0}p_{\alpha}(a_i|t_i^*)P_{\text{MD,M}}^{*1-y_i}(1-P_{\text{MD,M}}^*)^{y_i} \leq \max{(p_{e,1},p_{e,0})} < \\
    (\frac{1-p_{e,0}}{2})^{|\mathcal{M}|} \cdot \prod_{i:t_i=1}(1-p_{e,1})P_{\text{MD,L}}^{1-y_i}(1-P_{\text{MD,L}})^{y_i}.
    \end{aligned}
\end{equation}
Therefore, $(\boldsymbol{t},\hat{P}_{\text{MD,M}})$ results in a larger numerator than $(\boldsymbol{t}^*,P_{\text{MD,M}}^*)$ where $\boldsymbol{t}^*\neq \boldsymbol{t}$ and $\boldsymbol{a}=\boldsymbol{t}$, which means that $\boldsymbol{t}^*$ cannot be the maximizer. Hence, the event $\boldsymbol{t}^*\neq \boldsymbol{t}$ in this case has probability $0$, which concludes our proof.
\end{proof}
Now, we can state the main result of this section with the following proposition.
\begin{proposition}
Assume that \cref{assumption:bernouilli} holds. Let $(\boldsymbol{t_n}^*,P_{\text{MD,M}}^*)$ and $(\boldsymbol{t_d}^*,P_{\text{FA,M}}^*)$ be maximizers of the numerator and denominator in \eqref{eq:GLRT_ML_decision_independent}, respectively.
If $\mathbb{E}[a_i | t_i = 0] \to 0$ for every $i\in\mathcal{M}$ and \mbox{$\mathbb{E}[a_i | t_i = 1] \to 1$ for every $i\in\mathcal{L}$}, then, with high probability, the A-GLRT algorithm is equivalent to the likelihood ratio test using the measurements of legitimate robots only, that is
\begin{flalign}
\frac{\prod_{i\in\mathcal{L}} P_{\text{FA,L}}^{{y_i}}\cdot(1-P_{\text{FA,L}})^{1-{y_i}}}{\prod_{i\in\mathcal{L}}(1-P_{\text{MD,L}})^{{y_i}}\cdot P_{\text{MD,L}}^{1-{y_i}}}\underset{\mathcal{H}_{0}}{\overset{\mathcal{H}_{1}}{\gtrless}}\frac{\Pr(\Xi=0)}{\Pr(\Xi=1)}.
\label{eq:GLRT_legitimate}
\end{flalign}
\end{proposition}
\begin{proof}
By Lemma~\ref{lem:t_correctly_estimated}, we have $\boldsymbol{t_n}^*= \boldsymbol{t_d}^*=\boldsymbol{t}$ with high probability. We use $\boldsymbol{t}$ in place of both $\boldsymbol{t_n}^*$ and $\boldsymbol{t_d}^*$ for simplicity in the rest of the proof. First, in the trivial case where $\sum_{i:t_i=0}1=0$, the GLRT has the form \eqref{eq:GLRT_legitimate} because there are no malicious robots in the system. In other cases, $$P_{\text{MD,M}}^*=\frac{\sum_{i:t_i=0}(1-{y_i})}{\sum_{i:t_i=0}1},$$ by Lemma~\ref{lem:lemma_t_given}. Similarly, $$P_{\text{FA,M}}^*=\frac{\sum_{i:t_i=0}{y_i}}{\sum_{i:t_i=0}1}.$$ Notice that $P_{\text{MD,M}}^*$ equals $1-P_{\text{FA,M}}^*$. Therefore, in the calculation of GLRT, the contribution coming from the malicious robots in the numerator and denominator cancel each other out. As a result, the GLRT has the form \eqref{eq:GLRT_legitimate}. Therefore, in all cases, the GLRT has the form \eqref{eq:GLRT_legitimate} with high probability.
\end{proof}

\subsection{Utilizing the Prior Knowledge with A-GLRT} \label{subsec:additional_info-GLRT}
In this section, we introduce two different modifications of the A-GLRT algorithm to incorporate additional information about the malicious robots into the system. 
\subsubsection{Probability of Each Robot Being Malicious}
\label{subsubsec:prior-GLRT}
In some cases, the probability of each robot being malicious is available or assumed to be known. Essentially, this information would quantify the vulnerability of the multi-robot system, where a higher probability would correspond to a more vulnerable system. For instance, the previous works \cite{rawat2010collaborative,marano2008distributed} have this assumption. In this part, we modify the A-GLRT algorithm to introduce a way to use this additional information. First, we formalize this new assumption.
\begin{assumption}
\label{assumption:known_prior}
Let $t_i$ denote the true identity of a robot $i$ in the network. We assume that the prior distribution of robot $i$ being legitimate or malicious, denoted by $\Pr(t_i)$, is the same for all robots $i$ and it is independent of other robots. Moreover, we assume that these prior probabilities is known by the FC.
\end{assumption}
Under this assumption, we modify the GLRT given by \eqref{eq:GLRT_ML_decision_independent} as follows:
\begin{flalign}\label{eq:GLRT_known_prior}
\frac{\max\limits_{\boldsymbol{t}\in\{0,1\}^N,P_{\text{MD,M}}\in[0,1]}\Pr(\boldsymbol{a},\boldsymbol{t})\Pr(\boldsymbol{y}|\mathcal{H}_1,\boldsymbol{t},P_{\text{MD,M}})}
{\max\limits_{\boldsymbol{t}\in\{0,1\}^N,P_{\text{FA,M}}\in[0,1]}\Pr(\boldsymbol{a},\boldsymbol{t})\Pr(\boldsymbol{y}|\mathcal{H}_0,\boldsymbol{t},P_{\text{FA,M}})}
\underset{\mathcal{H}_{0}}{\overset{\mathcal{H}_{1}}{\gtrlesseqslant}} \gamma_{\text{GLRT}},
\end{flalign}
where we can calculate $\Pr(\boldsymbol{a},\boldsymbol{t})$ using $$\Pr(\boldsymbol{a},\boldsymbol{t})=\Pr(\boldsymbol{a}|\boldsymbol{t})\Pr(\boldsymbol{t}).$$
Now, we focus on how to calculate the numerator with this new formulation since the denominator follows a similar structure. We write the numerator as:
\begin{equation}
    \begin{aligned}
    \max_{\boldsymbol{t}\in\{0,1\}^N,P_{\text{MD,M}}\in[0,1]}\left\{\prod_{i:t_i=1}\Pr(a_i,t_i)P_{\text{MD,L}}^{1-y_i}(1-P_{\text{MD,L}})^{y_i} \right. \cdot \\ 
    \left. \prod_{i:t_i=0}\Pr(a_i,t_i)P_{\text{MD,M}}^{1-y_i}(1-P_{\text{MD,M}})^{y_i} \right\}.
    \end{aligned}
\end{equation}
Notice that this new formulation does not affect the results in Lemma~\ref{lem:lemma_t_given}. Moreover, let 
\begin{flalign}
    c_{\text{L},i}&\triangleq p_{\alpha}(a_i|t_i=1)\Pr(t_i=1)P_{\text{MD,L}}^{1-y_i}(1-P_{\text{MD,L}})^{y_i}, \nonumber\\
    c_{\text{M},i}(P_{\text{MD,M}})&\triangleq p_{\alpha}(a_i|t_i=0)\Pr(t_i=0)P_{\text{MD,M}}^{1-y_i}(1-P_{\text{MD,M}})^{y_i}.
\end{flalign} 
With these new definitions, Lemma~\ref{lem:lemma_P_given} and Theorem~\ref{thm:efficient_maximization} still hold. Therefore, we can still use the A-GLRT algorithm given in Algorithm~\ref{alg:A-GLRT} just by replacing $ c_{\text{L},i}$ and $ c_{\text{M},i}$ with these new definitions that include $\Pr(t_i)$.

\subsubsection{An Upper Bound on The Number of Malicious Robots}
\label{subsubsec:upper_bound-GLRT}
In this part, we assume that the the upper bound on the proportion of the malicious robots in the network, denoted by $\overline{m}$, is known similar to the Two-Stage Approach algorithm. Now, we show how to modify Algorithm~\ref{alg:A-GLRT} to incorporate this additional information. This upper bound can be expressed as $$|\mathcal{M}|\triangleq \sum_{i\in\mathcal{N}}1-t_i \leq \overline{m}N.$$ First, notice that this new constraint on $\boldsymbol{t}$ does not affect the results in Lemma~\ref{lem:lemma_t_given}. However, the inner maximization given in expression \eqref{eq:GLRT_numerator_pma_first} turns into a constrained optimization problem, that is
\begin{equation}
    \begin{aligned}\label{eq:GLRT_known_m_inner_max}
    \max_{\boldsymbol{t}\in\{0,1\}^N,|\mathcal{M}|\leq \overline{m}N}\left\{\prod_{i:t_i=1}p_{\alpha}(a_i|t_i)P_{\text{MD,L}}^{1-y_i}(1-P_{\text{MD,L}})^{y_i} \right. \cdot \\
   \left.\prod_{i:t_i=0}p_{\alpha}(a_i|t_i)P_{\text{MD,M}}^{1-y_i}(1-P_{\text{MD,M}})^{y_i}\right\},
    \end{aligned}
\end{equation}
for a given $P_{\text{MD,M}}$. We provide Algorithm~\ref{alg:inner_constrained_max} to calculate this maximization.
\begin{algorithm}[h!]
\caption{Input: $\mathbf{y}$, $\mathbf{a}$, $P_{\text{MD,M}}$, $P_{\text{MD,L}}$ $p_{\alpha}(a_i|t=1)$, $p_{\alpha}(a_i|t=0)$, N, $\overline{m}$ \\ Output: Estimate $\hat{\boldsymbol{t}}$}
\label{alg:inner_constrained_max}
\begin{algorithmic}[1]

\State Initialize $N\times 1$ vector $\boldsymbol{d}$ arbitrarily.

\For{$i$=1 to N}

\State Set $c_{\text{L},i}=p_{\alpha}(a_i|t_i=1)P_{\text{MD,L}}^{1-y_i}(1-P_{\text{MD,L}})^{y_i}$.

\State Set $c_{\text{M},i}=p_{\alpha}(a_i|t_i=0)P_{\text{MD,M}}^{(1-y_i)}(1-P_{\text{MD},\text{M}})^{y_i}$.

\State Set $d_i = c_{\text{M},i}-c_{\text{L},i}$.

\EndFor

\State Set $\Tilde{\boldsymbol{d}} = \text{Sorted($\boldsymbol{d}$)}$  \Comment{sorting is in descending order}

\State $count=0$

\ForAll{$\Tilde{d}_j \in \Tilde{\boldsymbol{d}}$}
\State Set $i$ as the corresponding index of $j$ in the unordered vector $\boldsymbol{d}$
 \If {$\Tilde{d}_j > 0$ and $count < \overline{m}N$}
    \State $\hat{t}_i=0$ 
    \State Set $count=count+1$
 \Else
    \State $\hat{t}_i=1$ 
\EndIf
\EndFor

\State Return $\hat{\boldsymbol{t}}$.

\end{algorithmic}
\end{algorithm}

The main difference of Algorithm~\ref{alg:inner_constrained_max} compared to the unconstrained inner maximization described in \cref{lem:lemma_P_given} is that it requires sorting. One can use a sorting algorithm which takes $\mathcal{O}(N\log{N})$ comparisons such as merge sort \cite{knuth1998art}. Notice that this additional computation increases the number of comparisons given in Lemma~\ref{lem:lemma_P_given} from $\mathcal{O}(N)$ to $\mathcal{O}(N\log{N})$.

\section{Hardware Experiment and Numerical Results}
\label{sec:Results}
\begin{figure}[t!]
    \centering
    \includegraphics[scale=0.35]{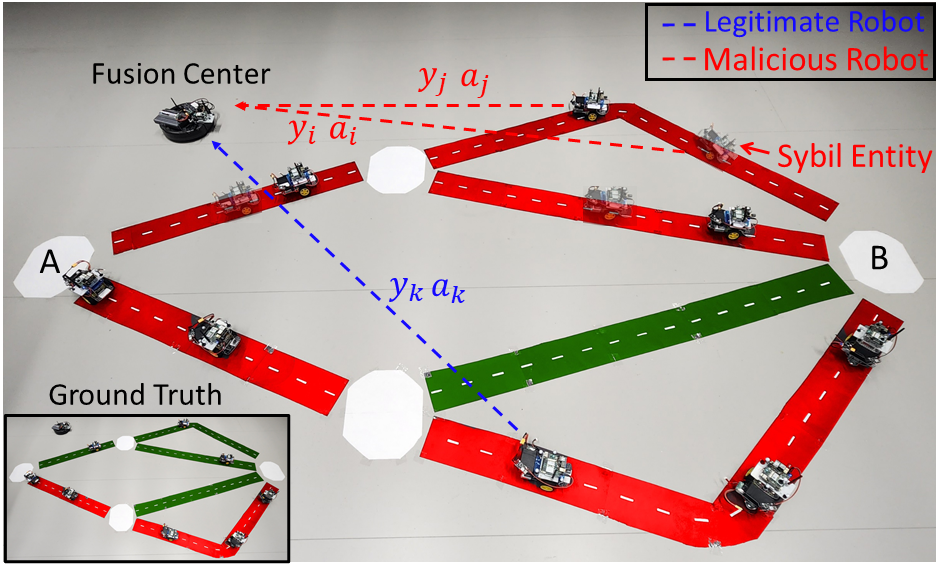}
    \caption[Experimental setup for testing the binary hypothesis testing algorithms.]{Robots drive along a roadmap comprised of six road segments to get from point A to point B. While traversing the roadmap, robots estimate the congestion on their current road segment as either containing traffic (red) or not (green), and relay their estimates to the FC. All robots relay messages to the FC, but only a few are depicted on the figure for ease of readability.}
    \label{fig:hardware_setup}
\end{figure}

We perform a hardware experiment with robotic vehicles driving on a mock-up road network where robots are tasked with reporting the traffic condition of their road segment to a FC. The objective of the malicious robots is to cause the FC to incorrectly perceive the traffic conditions (see Fig.~\ref{fig:hardware_setup}). A numerical study further demonstrates the performance of this scenario with an increasing proportion of malicious robots.

We compare the performance of the 2SA and A-GLRT against several benchmarks including the \emph{\textbf{Oracle}}, where the FC knows the true trust vector $\mathbf{t}$ and discards malicious measurements, (this serves as a lower bound on the probability of error), the \emph{\textbf{Oblivious FC}}, where the FC treats every robot as legitimate, and a \emph{\textbf{Baseline Approach}} \cite{rawat2010collaborative} where the FC uses a history of $T$ measurements to develop a reputation about each robot. The Baseline method ignores information from robots whose measurements disagree with the final decision at least $\eta < T$ times. The \emph{Oracle}, \emph{Oblivious FC}, and \emph{Baseline Approach} use the decision rule in \eqref{eq:detection_p_e_FC_legitimate_assumption}. Malicious robots perform a Sybil attack where they spoof additional robots into the network. We use the opensource toolbox in \cite{WSR_toolbox} to obtain trust values from communicated WiFi signals by analyzing the similarity between different fingerprints to detect spoofed transmissions. The works in \cite{yemini2021characterizing,AURO,CrowdVetting} model these trust values $\alpha_i \in [0,1]$ as a continuous random variable. We discretize the sample space by letting $\mathcal{A} = \{0,1\}$ and setting $a_i = 1$ if the measured trust value is $\geq 0.5$ and $a_i = 0$ otherwise.

\subsection{Hardware Experiment}

\begin{figure}[t!]
    \centering
    \includegraphics[scale=0.28]{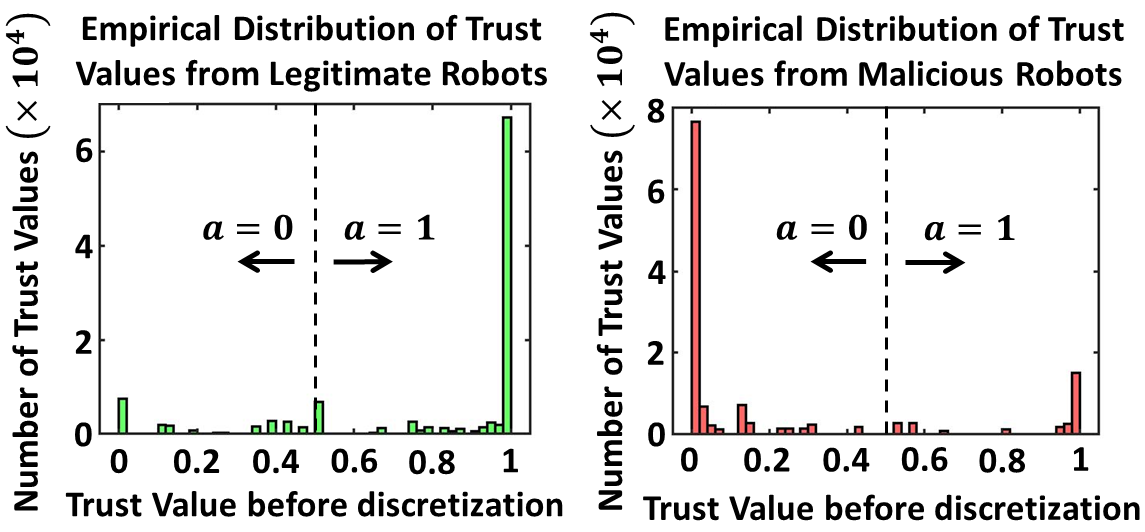}
    \caption[Empirical trust value results from the binary hypothesis testing experiments.]{Empirical distribution of the trust values gathered during the hardware experiment for legitimate and malicious robots. The trust value is thresholded to $a=1$ if it is $\geq 0.5$, and $a=0$ otherwise.}
    \label{fig:histograms}
\end{figure}

A group of $N = 11$ mobile robots drive in a loop from a starting point A to point B, approximately $4.5$ meters apart, by traversing one of four possible paths made up of six different road segments. As the robots drive between points A and B they are given noisy position information for themselves and neighboring robots from an OptiTrack motion capture system with added white Gaussian noise with a variance of $1m^2$. This serves as a proxy for GPS-reported measures used in crowdsourcing traffic estimation schemes like Waze, Google Maps, and others. A road segment is considered to have traffic ($y_i = 1$) if the number of robots on the segment is $\geq 2$. Of the $11$ robots in the group, $5$ robots are legitimate, $3$ are malicious, and $3$ are spoofed by the malicious robots (making them also malicious). Malicious robots know the true traffic conditions and report the wrong measurement with probability $0.99$, i.e., $P_{\text{FA,M}} = P_{\text{MD,M}} =0.99$. Hypothesis tests were run on each road segment any time at least one robot was present on that segment. The entire experiment was run for $15$ minutes with a frequency of $30$ hypothesis tests on each road segment per second. This led to a total of $61233$ hypothesis tests carried out. Of the $61233$ tests, $29.9 \%$ consisted of only legitimate robots, $28.1 \%$ of only malicious robots, and $42.0 \%$ contained both legitimate and malicious robots. The empirical data from the experiment is stated in \cref{tab:experiment}, where \emph{Baseline1} and \emph{Baseline5} refer to the Baseline Approach from \cite{rawat2010collaborative} with parameters $T$ and $\eta$ set to ($T = 1$, $\eta = 0.5$) and ($T = 5$, $\eta = 2.5$). We determined the parameters in \cref{tab:experiment} by first running an experiment without performing hypothesis tests and observing the behavior of the system compared to ground truth. The trust values gathered using the toolbox in \cite{WSR_toolbox} led to the empirical probabilities $p_{\alpha}(a_i=1|t_i=1) = 0.8350$ and $p_{\alpha}(a_i=1|t_i=0) = 0.1691$ (see \cref{fig:histograms}).

In our hardware experiment the 2SA and A-GLRT outperform the Oblivious FC and the Baseline Approach. The Baseline Approach exhibits a high percent error due to the fact that it relies on the majority of the network being legitimate. Since $6$ out of $11$ robots are malicious, it is likely that many hypothesis tests are conducted where the majority is malicious. This points to a common vulnerability of reputation based approaches that assume only a small proportion of the network is malicious.

\begin{table}[t!]
    \centering
    \begin{tabular}{|c|c|c|c|} \hline
    \multicolumn{4}{|c|}{\textbf{\textit{Parameters}}} \\ \hline
    $P_{\text{FA,L}}$  & 0.0800 & $P_{\text{MD,L}}$ & 0.2100 \\ \hline  $\Pr(\Xi=0)$  & 0.6432 & $\Pr(\Xi=1)$ & 0.3568 \\ \hline \multicolumn{4}{|c|}{\textbf{\textit{Percent Error}}} \\ \hline \textbf{2SA (Sec.~\ref{sec:2SA})} & \textbf{30.5} \% &\textbf{A-GLRT (Sec.~\ref{sec:GLRT})} & \textbf{29.0} \% \\ \hline Oracle & 19.5 \% & Oblivious FC & 52.0 \% \\ \hline Baseline1 & 50.8 \% & Baseline5 & 49.1 \% \\ \hline
    \end{tabular}
    \caption{Experimental Results}
    \label{tab:experiment}
\end{table}

\paragraph{\textbf{Numerical Study}}

\begin{figure}[t!]
    \centering
    \includegraphics[scale=0.44]{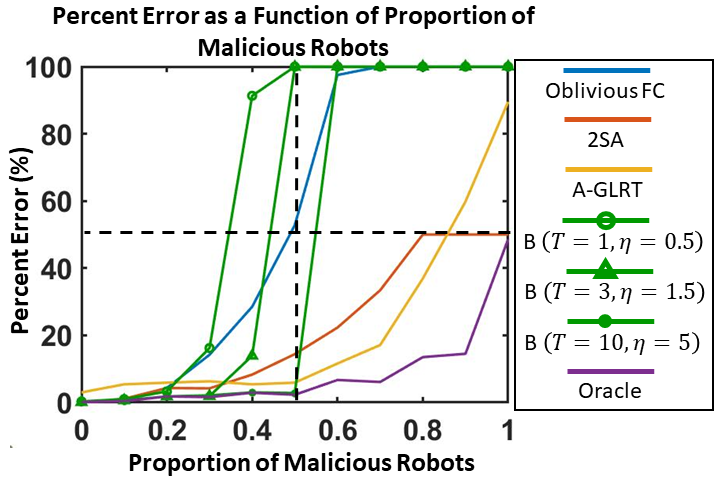}
    \caption[Numerical study comparing the performance of the Two Stage Approach and Adversarial Generalized Likelihood Ratio Test algorithms with an Oracle that has perfect knowledge of the robot identities and a baseline method that utilizes a history of sensor measurements to develop reputations for robots.]{The percent error for multiple hypothesis test approaches when the proportion of malicious robots is varied. The 2SA and A-GLRT outperform the Oblivious FC and Baseline Approaches (B) when the majority of the network is malicious. The performance of the Oracle declines as the proportion of malicious robots in the network increases since the FC is given access to less legitimate information.}
    \label{fig:sim_study}
\end{figure}

Next, we perform a numerical study on the performance of each approach when the proportion of malicious robots is varied. In the numerical study we use $N = 10$ robots with $\Pr(\Xi = 0) = \Pr(\Xi = 1) = 0.5$, $P_{\text{FA,L}} = P_{\text{MD,L}} = 0.15$, and $P_{\text{FA,M}} = P_{\text{MD,M}} = 0.99$ and perform hypothesis tests over $1000$ trials for each proportion of malicious robots. In the simulation study the trust value distributions are fixed at $p_{\alpha}(a_i = 1|t_i = 1) = 0.8$, $p_{\alpha}(a_i = 1|t_i = 0) = 0.2$, and the proportion of malicious robots varies from $0$ to $1$. The results of the simulation study are plotted in \cref{fig:sim_study}. From the plot it can be seen that the 2SA and the A-GLRT perform well even after the number of malicious robots exceeds majority since they use additional trust information independent of the data, whereas the Baseline Approaches (abbreviated with `B' in the figure) fail since they use only the data to assess the trustworthiness of the robots. Additionally, the existence of the critical proportion of malicious robots, $m^*$, beyond which the 2SA chooses to ignore all measurements and make the decision using the prior probabilities $\Pr(\Xi=0)$ and $\Pr(\Xi=1)$ can be seen. This value is approximately $m^* = 0.8$ for this set of parameters.

\addtolength{\textheight}{-4cm}   

\section{Conclusion}
\label{sec:conclusion}

In this paper we present two methods to utilize trust values in solving the binary adversarial hypothesis testing problem. The 2SA uses the trust values to determine which robots to trust, and then makes a decision from the measurements of the trusted robots. The A-GLRT jointly uses the trust values and measurements to estimate the trustworthiness of each robot, the strategy of malicious robots, and the true hypothesis.


\bibliographystyle{IEEEtran}
\bibliography{references.bib}

\end{document}